\def\year{2021}\relax
\documentclass[letterpaper]{article} 
\usepackage{aaai21}  
\usepackage{times}  
\usepackage{helvet} 
\usepackage{courier}  
\usepackage[hyphens]{url}  
\usepackage{graphicx} 
\usepackage{natbib}  
\usepackage{caption} 
\usepackage{booktabs}
\usepackage{color}
\usepackage{amsmath}
\usepackage{subcaption}
\usepackage{algorithm2e}
\usepackage{algpseudocode}
\usepackage{xcolor}
\usepackage{amsfonts}

\title{Efficient Poverty Mapping from High Resolution Remote Sensing Images}
\author {
    Kumar Ayush\thanks{Equal Contribution}\textsuperscript{\rm 1}
    Burak Uzkent\footnotemark[1]\textsuperscript{\rm 1}
    Kumar Tanmay\textsuperscript{\rm 3} 
    Marshall Burke\textsuperscript{\rm 2}
    David Lobell\textsuperscript{\rm 2}
    Stefano Ermon\textsuperscript{\rm 1} \\
}
\affiliations {
    \textsuperscript{\rm 1} Department of Computer Science, Stanford University \\
    \textsuperscript{\rm 2} Department of Earth Science, Stanford University \\
    \textsuperscript{\rm 3} Department of Electrical Engineering, IIT Kharagpur \\
    kayush@cs.stanford.edu, buzkent@cs.stanford.edu,
    kr.tanmay147@iitkgp.ac.in,
    mburke@stanford.edu, dlobell@stanford.edu, ermon@cs.stanford.edu
}
\begin{document}

\maketitle

\begin{abstract}

The combination of high-resolution satellite imagery and machine learning have proven useful in many sustainability-related tasks, including poverty prediction, infrastructure measurement, and forest monitoring. However, the accuracy afforded by high-resolution imagery comes at a cost, as such imagery is extremely expensive to purchase at scale. This creates a substantial hurdle to the efficient scaling and widespread adoption of high-resolution-based approaches.  To reduce acquisition costs while maintaining accuracy, we propose a reinforcement learning approach in which free low-resolution imagery is used to dynamically identify where to acquire costly high-resolution images, prior to performing a deep learning task on the high-resolution images. 
We apply this approach to the task of poverty prediction in Uganda, building on an earlier approach that used object detection to count objects and use these counts to predict poverty. Our approach exceeds previous performance benchmarks on this task while using 80\% fewer high-resolution images, and could be useful in many 
domains that require high-resolution imagery.
\end{abstract}

\section{Introduction}
When combined with machine learning, high-resolution satellite imagery has proven broadly useful for a range of sustainability-related tasks,
from poverty prediction \cite{jean2016combining,ayush2020generating,sheehan2019predicting,blumenstock2015predicting,yeh2020using} to infrastructure measurement \cite{cadamuro2018assigning} to forest and water quality monitoring \cite{fisher2018impact} to the mapping of informal settlements \cite{mahabir2018critical}. Compared to coarser (10-30m) publicly-available imagery~\cite{drusch2012sentinel2}, high-resolution ($<1m$) imagery has proven particularly useful for these tasks because it is often able to resolve specific objects or features that are 
undetectable in coarser imagery.

When combined with machine learning, high-resolution satellite imagery has proven broadly useful for object detection~\cite{lam2018xview}, object tracking~\cite{uzkent2018tracking,uzkent2017aerial}, cloud removal~\cite{sarukkai2020cloud}, and a range of sustainability-related tasks,
from poverty prediction \cite{jean2016combining,ayush2020generating,sheehan2019predicting,blumenstock2015predicting,yeh2020using} to infrastructure measurement \cite{cadamuro2018assigning}. Compared to coarser (10-30m) publicly-available imagery~\cite{drusch2012sentinel2}, high-resolution ($<1m$) imagery has proven particularly useful for these tasks because it is often able to resolve specific objects or features that are 
undetectable in coarser imagery.

For example, recent work demonstrated an approach for predicting local-level consumption expenditure using object detection on high-resolution daytime satellite imagery~\cite{ayush2020generating}, showing how this approach can yield interpretable predictions and also outperform previous benchmarks that rely on lower-resolution, publicly-available satellite imagery~\cite{drusch2012sentinel2}. This additional information, however, typically comes at a cost, as high-resolution satellite imagery must be purchased from private providers.  Additionally, processing high-resolution images is computationally more expensive than the coarser resolution ones
\cite{uzkent2019learning,zhu2016traffic,meng2017detecting,lampert2008beyond,wojek2008sliding,redmon2017yolov2,gao2018dynamic}. Given these costs, deploying these models at scale using high-resolution imagery quickly becomes cost-prohibitive for most organizations and research teams, inhibiting the broader development and deployment of machine-learning based tools and insights based on these data. 

To address this problem, we propose a reinforcement learning approach that uses coarse, freely-available public imagery to dynamically identify where to acquire costly high-resolution images, prior to conducting an object detection task.  This concept leverages publicly available Sentinel-2~\cite{drusch2012sentinel2} images (10-30m) to sample smaller amount of high-resolution images ($<$1m). Our framework is inspired from the recent studies in computer vision literature that perform conditional inference to reduce computational complexity of convolutional networks in test time~\cite{uzkent2020learning,wu2018blockdrop}.


We apply our approach to the domain of poverty prediction, and show how our approach can substantially reduce the cost of previous methods that 
used deep learning on high-resolution images to predict poverty~\cite{ayush2020generating} while maintaining or even improving their accuracy. In our study country of Uganda, we show how our approach can reduce the number of high-resolution images needed by 80\%, in turn reducing the cost of making a country-wide poverty map using this approach by an estimate \$2.9 million. 
\section{Poverty Mapping from Remote Sensing Imagery}
Poverty is typically measured using consumption expenditure, the value of all the goods and services consumed by a household in a given period. A  household or individual is said to be poverty stricken if their measured consumption expenditure falls below a defined threshold (currently \$1.90 per capita per day). We focus on this consumption expenditure as our outcome of interest, using ``poverty" as shorthand for ``consumption expenditure" throughout the paper. While typical household surveys measure consumption expenditure at the household level, publicly available data typically only release geo-coordinate information at the ``cluster" level -- which is a village in rural areas and a neighborhood in urban areas.  Efforts to predict poverty have thus focused on predicting at the cluster level (or more aggregated levels) \cite{ayush2020generating}. 

\citet{ayush2020generating} demonstrated state-of-the-art results for predicting village-level poverty using high-resolution satellite imagery, and showed how such predictions could be made with an interpretable model. In particular, they trained an object detector to obtain classwise object counts (buildings, trucks, passenger vehicles, railway vehicles, etc.) in high-resolution images, and then used these counts in a regression model to predict poverty. Not only were these categorical features predictive of poverty, but their counts had clear and intuitive relationships with the outcome of interest. The cost of this accuracy and interpretability was the high-resolution imagery, which typically must be purchased for \$10-20 per km$^2$ from private providers.

\textbf{Problem statement.} 
Let $\{(\mathcal{H}_i, \mathcal{L}_i, y_i, c_i)\}_{i=1}^N$ be a set of $N$ villages surveyed, where  $c_i = (c^{lat}_{i}, c^{lon}_{i})$ is the latitude and longitude coordinates for cluster $i$, and $y_{i} \in \mathbb{R}$is the corresponding average poverty index for a particular year. For each cluster $i$, we can acquire both high-resolution (at a cost) and low-resolution (free of charge) satellite imagery corresponding to the survey year, $\mathcal{H}_i \in \mathbb{R}^{W \times H \times B}$, a $W \times H$ image with $B$ channels, and $\mathcal{L}_i \in \mathbb{R}^{W/D \times H/D \times B}$, a $W/D \times H/D$ image with $B$ channels. Here $D$ represents a scalar to show the resolution difference between low-resolution and high-resolution images. 
Our goal is to learn (1) a regressor $f_r$ to predict the poverty index $y_i$ using $\mathcal{L}_i$ and parts of $\mathcal{H}_i$ (the informative regions) selected by (2) an adaptive data acquisition scheme based on $\mathcal{L}_i$. This adaptive data acquisition scheme is optimized to minimize cost (which depends on the number of selected regions) while maximizing the accuracy of $f_r$. 


\begin{figure*}[!h]
\centering
\includegraphics[width=0.8\textwidth]{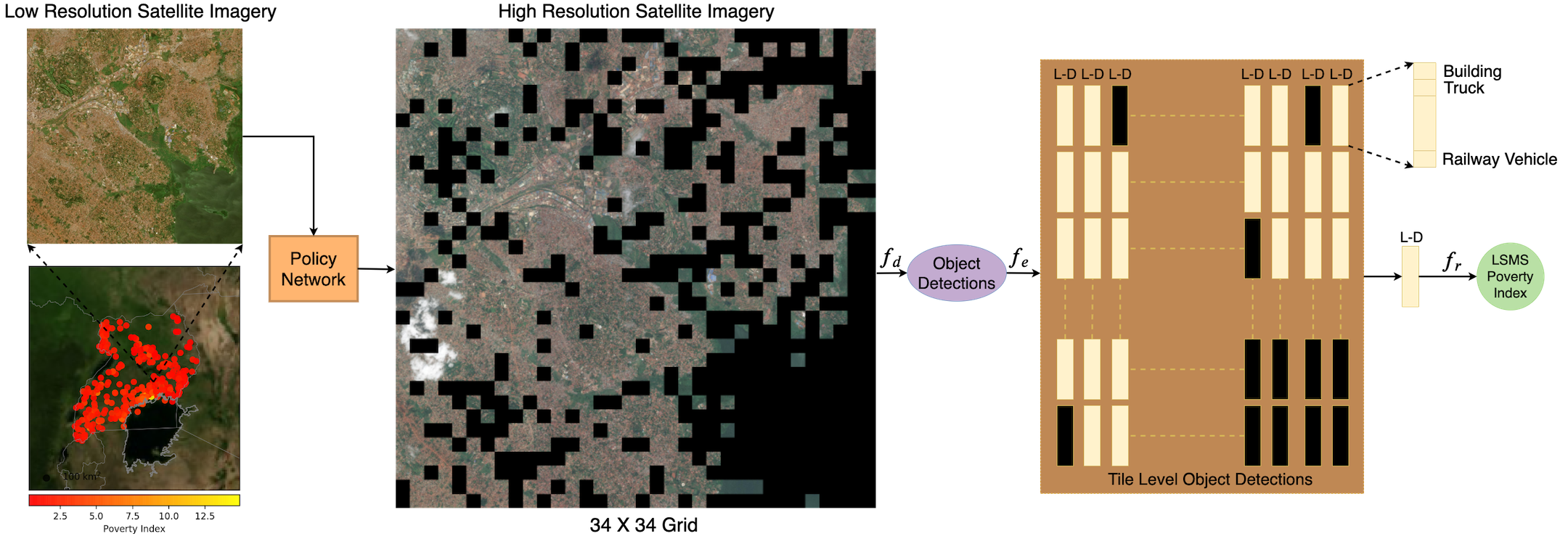}
\caption{Schematic overview of the proposed approach. The Policy Network uses cheaply available Sentinel-2 low-resolution image representing a cluster to output a set of actions representing unique 1000$\times$1000 px high-resolution tiles in the 34$\times$34 grid. Then object detection is performed on the sampled HR tiles (black regions represent dropped tiles) to obtain the corresponding class-wise object counts ($L$-dimensional vectors). Finally, the classwise object counts vectors corresponding to the acquired HR tiles are added element-wise to get the final feature vector representing the cluster. Our reinforcement learning approach dynamically identifies where to acquire high-resolution images, conditioned on cheap, low-resolution data, before performing object detection, whereas the previous work \cite{ayush2020generating} exhaustively uses all the HR tiles representing a cluster for poverty mapping, making their method expensive and less practical.
}
\label{fig:pipeline}
\end{figure*}

\section{Dataset}
\label{ugandaImages}
\textbf{Socio-economic Data.} Our ground truth dataset consists of data on consumption expenditure (poverty) from Living Standards Measurement Study (LSMS) survey conducted in Uganda by the Uganda Bureau of Statistics between 2011 and 2012 \cite{lsms}. The survey  consists of data from 2,716 households in Uganda, grouped into unique locations called clusters. The latitude and longitude, $c_i = (c^{lat}_i, c^{long}_i)$, of a cluster $i=\{1, 2, \dots, N\}$ is given, with noise of up to $5$ km added in each direction by the surveyers to protect respondent privacy. Individual household locations in each cluster $i$ are also withheld to preserve anonymity. We have N=320 clusters in the survey which we use to test the method performance in terms of predicting the average poverty index, $y_i$, for a group $i$. For each $c_i$,
the survey measures the poverty level by the per capital daily consumption in dollars which we refer to as the ``LSMS poverty score" for simplicity like \cite{ayush2020generating}. Fig. \ref{fig:pipeline} (bottom left corner) visualizes the surveyed locations on the map along with their corresponding LSMS poverty scores, revealing that a high percentage of surveyed locations have relatively low consumption expenditure values. 

\textbf{Satellite Imagery.} We acquire both high-resolution and low-resolution satellite imagery for Uganda. The high-resolution satellite imagery, $\mathcal{H}_{i}$, corresponding to cluster $c_i$ (roughly, a village or neighborhood) is represented by T=34$\times$34=1156 images of 1000$\times$1000 pixels each with 3 channels, arranged in a 34$\times$34 square grid. This corresponds to a 10km$\times$10km spatial neighborhood centered at $c_i$. A large neighborhood is considered to deal with up-to 5km of random noise in the cluster coordinates that has been added by the survey organization to protect respondent privacy. These high-resolution images come from DigitalGlobe satellites with 3 bands (RGB) and 30cm pixel resolution. Formally, we represent all the high-resolution images corresponding to $c_i$ as a sequence of $T$ tiles as $\mathcal{H}_{i} = \{H_{i}^{j}\}_{j=1}^T$. We acquire all the high-resolution tiles representing a cluster for comparison with~\cite{ayush2020generating}. However, in real-word scenario our method requires only a small fraction of HR tiles in test time unlike~\cite{ayush2020generating} that acquires HR tiles exhaustively.

We also acquire low-resolution satellite imagery, $\mathcal{L}_{i}$, corresponding to cluster $c_i$ and represented by a single image of $1014 \times 1014$ pixels with $3$ channels. These images come from Sentinel-2 with 3 bands (RGB) and 10m pixel resolution and are freely available to the public. Each image corresponds to the same 10km$\times$10km spatial neighborhood centered at $c_i$, however the resolution is much lower -- each Sentinel-2 pixel corresponds to roughly 1000 pixels from the high-resolution imagery. Because of this low-resolution, it is not possible to perform fine-grained object detection just using these images. 
Fig. \ref{fig:pipeline} illustrates an example cluster from Uganda.


\section{Fine-grained Object Detection on High-Resolution Satellite Imagery}
Similar to \cite{ayush2020generating}, we use an intermediate object detection phase to obtain categorical features (classwise object counts) from high-resolution tiles of a cluster. Due to lack of object annotations for satellite images from Uganda, we use the same transfer learning strategy as in \cite{ayush2020generating} by training an object detector (YOLOv3~\cite{redmon2018yolov3}) on 
xView~\cite{lam2018xview}, one of the largest and most diverse publicly available overhead imagery datasets for object detection with $10$ parent-level and $60$ child-level classes. Earlier work \cite{ayush2020generating} studied both parent-level and child-level detectors and empirically find that not only the parent-level object detection features are better for poverty regression but at the same time are more suited for interpretability due to household level descriptions. Thus, we train YOLOv3 detector using parent-level classes (see x-axis labels of Fig. \ref{fig:counts_diff}).

As described in previous section
, each $\mathcal{H}_i$ representing a cluster is a set of $T$ high-resolution images, $\{H_i^j\}_{j=1}^T$. To obtain a baseline model that uses all the high-resolution imagery available, we follow the protocol in \cite{ayush2020generating} and run the trained YOLOv3 object detector on each 1000$\times$1000px tile (\emph{i.e.} $H_i^j$) to get the correspoding set of object detections 
. 
Similar to \cite{ayush2020generating}, we use these object detections to generate a $L$-dimensional vector, $\mathbf{v}_i^j \in \mathbb{R}^L$ (where $L$=10 is the number of object labels/classes), by counting the number of detected objects in each class. 
This class-wise object counts can be  used in a regression model for poverty estimation \cite{ayush2020generating}. 

\citet{ayush2020generating} exhaustively uses all T=1156 HR tiles of a cluster for poverty estimation. In contrast, we propose to use a method that adaptively selects informative regions for high-resolution acquisition conditioned on the publicly available, low-resolution data. Thus, we reduce the dependency on HR images that are expensive to acquire thereby reducing the costs of poverty prediction models that use HR images exhaustively~\cite{ayush2020generating} making their method costly and less practical.
We describe our solution in the next section.


\section{Adaptive Tile Selection}
\label{sec:generic}

Due to the large acquisition cost of HR images, it is non-trivial and expensive to deploy models based on HR imagery at scale. For this reason, we propose an efficient tile selection framework to capture relevant fine level information such as classwise object counts for downstream tasks. We represent the HR image covering a spatial cluster $i$ centered at $c_{i}=(c_{i}^{lat},c_{i}^{lon})$ as $\mathcal{H}_{i} \in R^{W \times H \times B}$ where $W$, $H$ and $B$ represent height width and number of bands. Additionally, we represent the LR image of the same spatial cluster $i$ as $\mathcal{L}_{i} \in R^{W/D, H/D, B}$ where $D$ represents a scalar for the number of pixels in width and height. For example, in the case of Sentinel-2 (10 m GSD), we have $D=30$ times smaller number of pixels than the high-resolution DigitalGlobe images (0.3m GSD).
With an adaptive approach, our task is to acquire only small subset of $\mathcal{H}_{i}$ conditionally on $\mathcal{L}_{i}$ while not hurting the performance in our downstream tasks that uses object counts from the cluster $i$.
This adaptive method is formulated as a two-step episodic Markov Decision Process (MDP), similar to~\cite{uzkent2020efficient}. In the first step, we adaptively sample HR tiles and in the second step, we run them through a pre-trained detector.


\textbf{Task Definition.}
The first module of our framework finds HR tiles to sample/acquire, conditioned on the low spatial resolution image covering a cluster (which is always acquired). However, a cluster is represented by 34000$\times$34000 px HR images. Directly learning actions with reinforcement learning on such a large area can be very challenging and unstable. For this reason, we decompose our task to many independent sub-tasks where each sub-task focuses on sampling the important parts of the corresponding area with HR images. Following this, we divide a cluster-level HR image $\mathcal{H}_{i} = (H_{i}^{1}, H_{i}^{2},\dotsc, H_{i}^{T})$ into equal-size non-overlapping tiles, where $T$ is the number of tiles. Similar to $\mathcal{H}_{i}$, we decompose 
$\mathcal{L}_{i}$ as $\mathcal{L}_{i} = (l_{i}^{1}, l_{i}^{2},\dotsc, l_{i}^{T})$ where $l_{i}^{j}$ represents the lower spatial resolution version (from Sentinel-2) of $H_{i}^{j}$. In this set up, we model $\mathcal{H}_{i}$ as a latent variable as it is not directly observed and it is inferred from the observation $\mathcal{L}_{i}$. We associate each tile, $H_{i}^{j}$, of $\mathcal{H}_{i}$ with an $L$-dimensional classwise object counts feature represented as $\mathbf{v}_{i}^{j}$.

In a simple scenario, we can take a single binary action for each $H_{i}^{j}$ whether to acquire it or not conditioned on $l_{i}^{j}$. However, we believe that choosing multiple actions representing different disjoint subtiles of tile $H_{i}^{j}$ can help us avoid sampling areas of tile $H_{i}^{j}$ where there are no objects of interest. 
For this reason, we divide tile $H_{i}^{j}$ into $S$ number of disjoint subtiles as $H_{i}^{j}=(h_{i}^{j,1},h_{i}^{j,2},...,h_{i}^{j,S})$. We then define our task as learning a policy network conditioned on $l_{i}^{j}$ to only choose HR sub-tiles from $H_{i}^{j}$ where there is desirable number of objects characterized by a reward function. Once we learn the policy network, in test time we run it on each $l_{i}^{j}$ of a cluster $i$ to sample HR images and run them through detector to find out the cluster-level object counts.

\textbf{1st Step of MDP.} In the first step, the agent observes $l_{i}^{j}$ and outputs a binary action array, $\mathbf{a}_{i}^{j} \in \{0,1\}^{S}$, where $a_{i}^{j,k} = 1$ represents acquisition of the HR version of the $k$-th subtile of $H_{i}^{j}$ \emph{i.e.} $h_{i}^{j,k}$. The subtile sampling policy, parameterized by $\theta_{p}$, is formulated as
    $
    \pi(\mathbf{a}_{i}^{j} | l_{i}^{j};\theta_{p}) = p(\mathbf{a}_{i}^{j}|l_{i}^{j};\theta_{p})
$
where $\pi(l_{i}^{j};\theta_{p})$ is a function mapping the observed LR image to a probability distribution over subtile sampling actions $\mathbf{a}_i^j$. 


\textbf{2nd Step of MDP.} In the second step, the agent runs the object detection on the selected HR subtiles. Conditioned on $\mathbf{a}_{i}^{j}$, it observes HR subtiles if necessary and produces $\mathbf{\hat{v}}_{i}^j$, a $L$-dimensional classwise object counts vector. We find the object counts with our adaptive framework using a pre-trained object detector $f_d$ (parameterized by $\theta_d$) as:
\begin{equation}
\hat{\mathbf{v}}_{i}^{j,k} = \left\{ \begin{array}{ll} f_{d}(h_{i}^{j,k}) \quad \text{if} \quad a_{i}^{j,k}=1 \\ \mathbf{0} \quad \text{else} \end{array} \right.
\end{equation}
Then, we compute the tile level object counts as $\mathbf{\hat{v}}_{i}^j = \sum_{k=1}^{S} \mathbf{\hat{v}}_{i}^{j,k}$. Finally,  we define our overall cost function $J$ as:
\begin{equation}
\max_{\theta_{p}} J(\theta_{p}, \theta_{d}) = \mathbb{E}_p[R(\mathbf{a}_i^j, {\mathbf{\hat{v}}_i^j}, \mathbf{v}_i^j)],
\label{Eq:Cost_Function}
\end{equation}
where the reward depends on $\mathbf{a}_{i}^{j}$, $\mathbf{\hat{v}}_i^j$, $\mathbf{v}_i^j$. 
Our goal is to learn the parameters $\theta_{p}$ given a pre-trained object detector $\theta_d$ to maximize the objective being a function of the reward function. 

\textbf{The Reward Function.} 
The desired outcome from our adaptive strategy is to reduce the \emph{image acquisition cost} drastically by sampling smaller subset of tiles. Taking this into account, we design a dual reward function that encourages dropping as many subtiles as possible while successfully approximating the classwise object counts. We define $R$ as follows:

\begin{equation}
R = R_{acc}(\mathbf{\hat{v}}_i^j, \mathbf{v}_i^j) + R_{cost}(\mathbf{a}_{i}^{j}) 
\end{equation}
\begin{equation}
R_{acc} = -||\mathbf{v}_i^j - \hat{\mathbf{v}}_i^j||_1
\end{equation}
\begin{equation}
R_{cost} = \lambda(1 - ||\mathbf{a}_{i}^{j}||_1 / S)
\label{eq:tradeoff}
\end{equation}
where $R_{acc}$ is object counts approximation accuracy and $R_{cost}$ represents the image acquisition cost with $\lambda$ as its coefficient. The $R_{acc}$ term encourages acquiring a subtile when the counts difference between the object counts from fixed HR subtile sampling policy and the adaptive policy is positive. 
We increase the reward \emph{linearly} with the smaller number of acquired subtiles for the cost component. 

\section{Modeling and Optimization of the Policy Network}
\label{sec:reward_function}
In the previous section, in high level we formulated the task of efficient HR subtile selection as a two step episodic MDP. In this section, we model how to learn the policy distribution for subtile sampling.

\textbf{Modeling the Policy Network.} In this study, we have $T=1156$ number of tiles as we have a 34$\times$34 grid of images. In this case, each grid consists of 2000$\times$2000 pixels. As mentioned in the previous section, we divide each tile into S=4 subtiles of 1000$\times$1000 pixels each (higher values of S led to unstable training with higher variance and less sparse selections).
In this study, similar to~\cite{uzkent2020efficient} we model the action likelihood function of the policy network, $f_{p}$, using the product of bernoulli distributions as:
\begin{align}
  \pi(\mathbf{a}_{i}^{j}|l_{i}^{j}; \theta_{p}) &= \prod_{k=1}^{S} (s_{i}^{j,k})^{a_i^{j,k}} (1- s_{i}^{j,k})^{(1-a_{i}^{j,k})} 
\label{eq:action_likelihood}
\end{align}
\begin{align}
    s_{i}^{j} &= f_{p}(l_{i}^{j};\theta_{p}) 
\label{eq:policy_network}
\end{align}
We use a sigmoid function to transform logits to probabilistic values, $s_{i}^{j,k} \in [0,1]$.


\textbf{Optimization of the Policy Network.} 
The previously defined objective function as shown in Eq.~\ref{Eq:Cost_Function} is not differentiable w.r.t the policy network parameters, $\theta_{p}$, because acquistion actions are discrete. 
To overcome this, we train using Policy Gradient~\cite{sutton2018reinforcement}. Our final objective function as shown below includes the reward function as well as action likelihood distribution which can be differentiated w.r.t $\theta_p$.
\begin{equation}
\nabla_{\theta_{p}}J = \mathbb{E}\left[ R(\mathbf{a}_{i}^{j}, \mathbf{\hat{v}}_{i}^{j}, \mathbf{v}_i^{j})\nabla_{\theta_{p}} \log \pi_{\theta_{p}}(\mathbf{a}_{i}^{j}|l_{i}^{j}) \right],
\label{eq:policy_gradient_coarse} 
\end{equation}
Our objective function relies on mini-batch Monte-Carlo sampling to approximate the expectation. Especially, in scenarios where we can not afford large mini-batches, we can have highly oscillating expectations which results in large variance. As this can de-stabilize the optimization, we use the self-critical baseline~\cite{rennie2017self}, $A$, to reduce the variance.
\begin{equation}
\nabla_{\theta_{p}}J = \mathbb{E}\left[ A \sum_{k=1}^{S} \nabla_{\theta_{p}}\log(s_{i}^{j,k}\mathbf{a}_{i}^{j,k}+(1-s_{i}^{j,k})(1-\mathbf{a}_{i}^{j,k})) \right]
\end{equation}
\begin{equation}
    A(\mathbf{a}_{i}^{j}, \mathbf{a}_{i}^{\prime j}) = R(\mathbf{a}_{i}^{j}, \mathbf{\hat{v}}_{i}^{j}, \mathbf{v}_{i}^{j}) - R(\mathbf{a}_{i}^{\prime j}, \mathbf{\hat{v}}_{i}^{\prime j}, \mathbf{v}_{i}^{j})
\end{equation}

where $\mathbf{a}_{i}^{\prime j}$ represents the baseline action vector. To get $\mathbf{a}_{i}^{\prime j}$, we use the most likely action vector proposed by the policy network: \textit{i.e.}, $a_{i}^{\prime j,k}=1$ if $s_{i}^{j,k}>0.5$ and $a_{i}^{\prime j,k}=0$ otherwise. 
Finally, in this study we use temperature scaling~\cite{sutton2018reinforcement} to adjust exploration/exploitation trade-off during optimization time as
\begin{equation}
s_{i}^{j,k} = \alpha s_{i}^{j,k} + (1-\alpha)(1-s_{i}^{j,k}).
\end{equation}
Setting $\alpha$ to a large value results in sampling from the learned policy whereas the small values lead to sampling from random policy. See appendix for the pseudocode and implementation details.

\section{Experiments}

\subsection{Training and Testing the Policy Network on xView}
Our goal is to learn policies to reduce the dependency on HR images in approximating object counts in a geocluster while successfully predicting the downstream index (poverty prediction). Since our downstream dataset (Uganda) does not contain object bounding boxes, it is not possible to assess how well we approximate true object counts. To achieve this, we train our policy network on the xView dataset where our object detector is trained on. We use $2000\times2000$ px images and their corresponding $224\times224$ px LR images to train the policy network on each point. As proposed earlier, the action space has 4 units representing the top left, top right, bottom left, and bottom right part ($1000\times1000$ px) of the full area. The detector is only run on the part chosen by the policy network. We train the policy network on 1249 points and test it on 200 points and show the results in Table~\ref{tab:xview_exps}.

Our policy network uses $42.3\%$ HR images while approximating the fixed approach in mean Average Precision (mAP) and mean Average Recall (mAR) metrics~\cite{redmon2018yolov3}.
This results indicate that the policy network learns to successfully choose regions where there are objects of interest and eliminate the regions with no objects of interest. See the Appendix for more details.

\begin{table}[!htb]
\centering
\resizebox{0.40\textwidth}{!}{
\begin{tabular}{@{}lllll@{}}
\toprule
 & \textbf{mAP} & \textbf{mAR} & \textbf{HR} & \textbf{Run-time} \\ \midrule
\textbf{No Dropping} & 24.3$\%$ & 42.5$\%$ & 100.0$\%$ &  2890 ms \\
\textbf{RL Method} & 26.3$\%$ & 41.1$\%$ & 42.3$\%$ &  1510 ms\\ \bottomrule
\end{tabular}}
\caption{Results on the xView test set.}
\label{tab:xview_exps}
\end{table}
\begin{table*}[!h]
\centering
\resizebox{\textwidth}{!}{
\begin{tabular}{lllllllllll}
    \toprule
    {} & \textbf{No Dropping} & \textbf{Fixed-18} & \textbf{Random-25}&\textbf{Stochastic-25}&\textbf{Green}&\textbf{Counts Pred.}&\textbf{Sett. Layer}&\textbf{Nightlights}&\textbf{Ours (Dry sea.)}&\textbf{Ours (Wet sea.)}\\
    \midrule
    \textbf{$r^2$}      & 0.53 & 0.43 & 0.34 & 0.26 & 0.33 & 0.49 & 0.45 & 0.45 & 0.51 $\pm$ 0.01 &\textbf{0.61 $\pm$ 0.01}\\
    MSE                 & 1.86 & 2.20 & 2.67 & 3.13 & 2.56 & 1.91 & 2.16 & 2.17 & 1.89 $\pm$ 0.02 &\textbf{1.46 $\pm$ 0.02}\\
    Explained Variance  & 0.54 & 0.43 & 0.33 & 0.27 & 0.36 & 0.48  & 0.46 & 0.45 & 0.50 $\pm$ 0.01 &\textbf{0.63 $\pm$ 0.02}\\
    HR Acquisition.     & 1.0  & 0.18 & 0.25 & 0.25 & 0.19 & 0.19 & 0.19 & 0.12 & 0.19 &\textbf{0.19}\\
    \bottomrule
  \end{tabular}}
\caption{LSMS poverty score prediction results in Pearson's $r^2$ (and two other metrics) for various methods. \emph{HR Acquisition} represents the fraction of HR tiles acquired. We report the mean and std of our RL model across 7 runs with different seeds.}
\label{tab:comparison}
\end{table*}

\begin{figure*}[!htb]
\centering
\begin{subfigure}[b]{0.18\textwidth}
    \includegraphics[width=\textwidth]{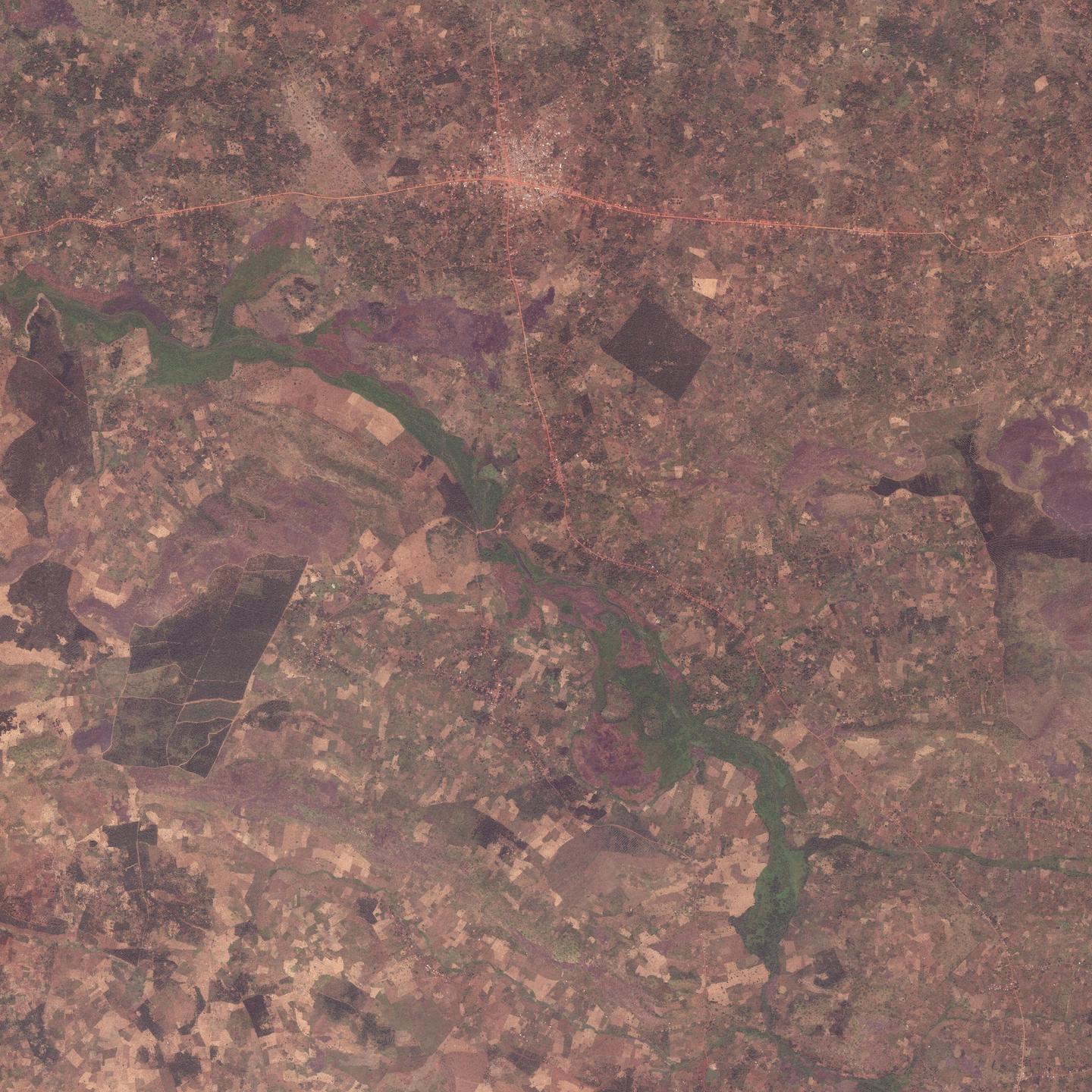}
    \caption{}
    \label{fig:original_dg1}
\end{subfigure}%
~
\begin{subfigure}[b]{0.18\textwidth}
    \includegraphics[width=\textwidth]{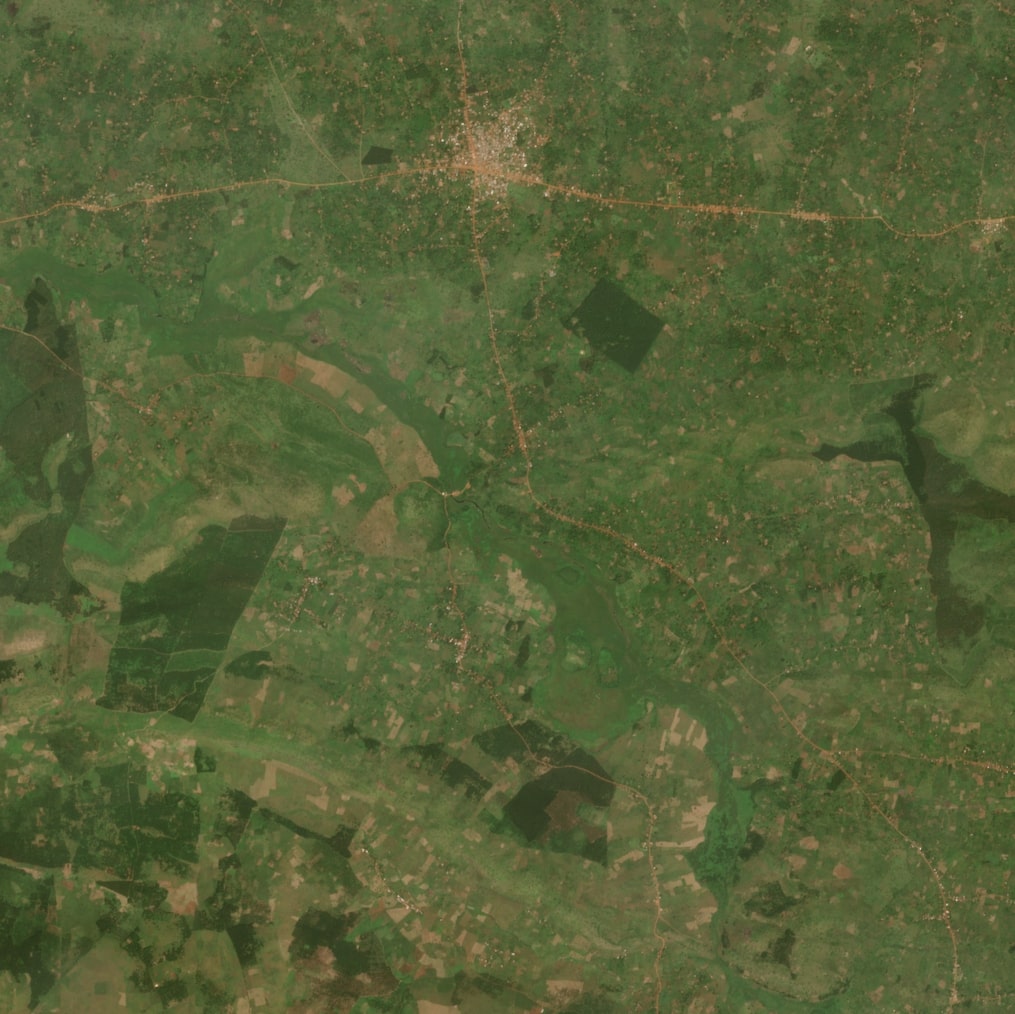}
    \caption{}
    \label{fig:dry_sentinel1}
\end{subfigure}%
~
\begin{subfigure}[b]{0.18\textwidth}
    \includegraphics[width=\textwidth]{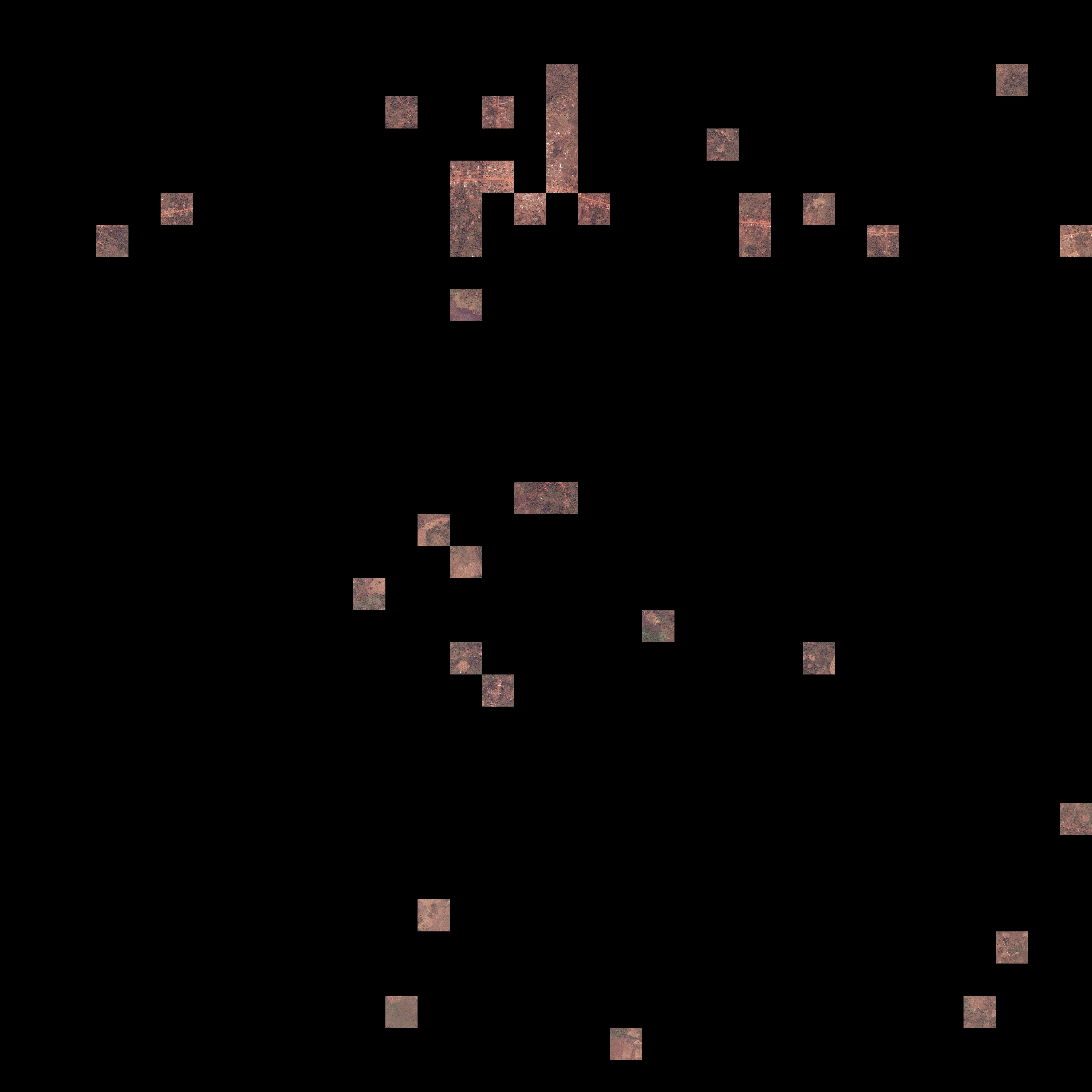}
    \caption{}
    \label{fig:dry_pred_sentinel1}
\end{subfigure}%
~
\begin{subfigure}[b]{0.18\textwidth}
    \includegraphics[width=\textwidth]{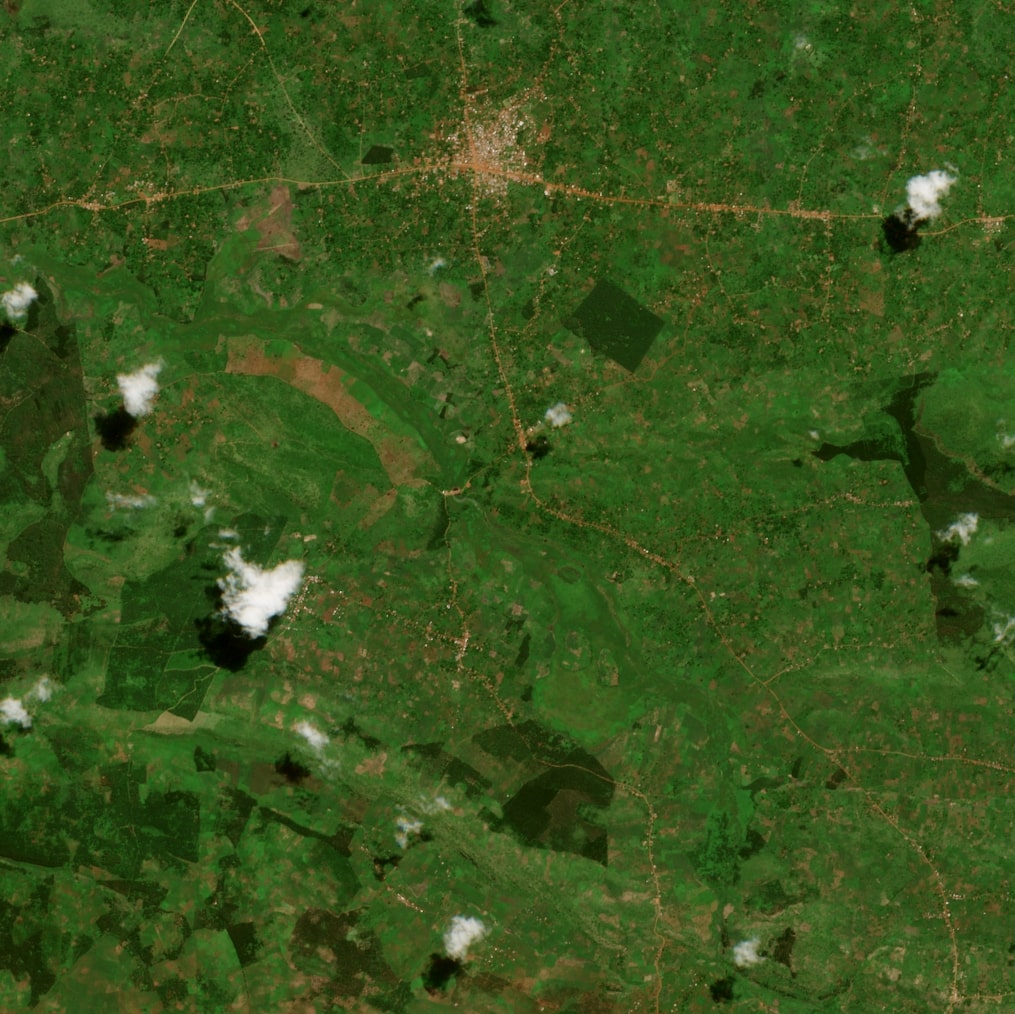}
    \caption{}
    \label{fig:wet_sentinel1}
\end{subfigure}%
~
\begin{subfigure}[b]{0.18\textwidth}
    \includegraphics[width=\textwidth]{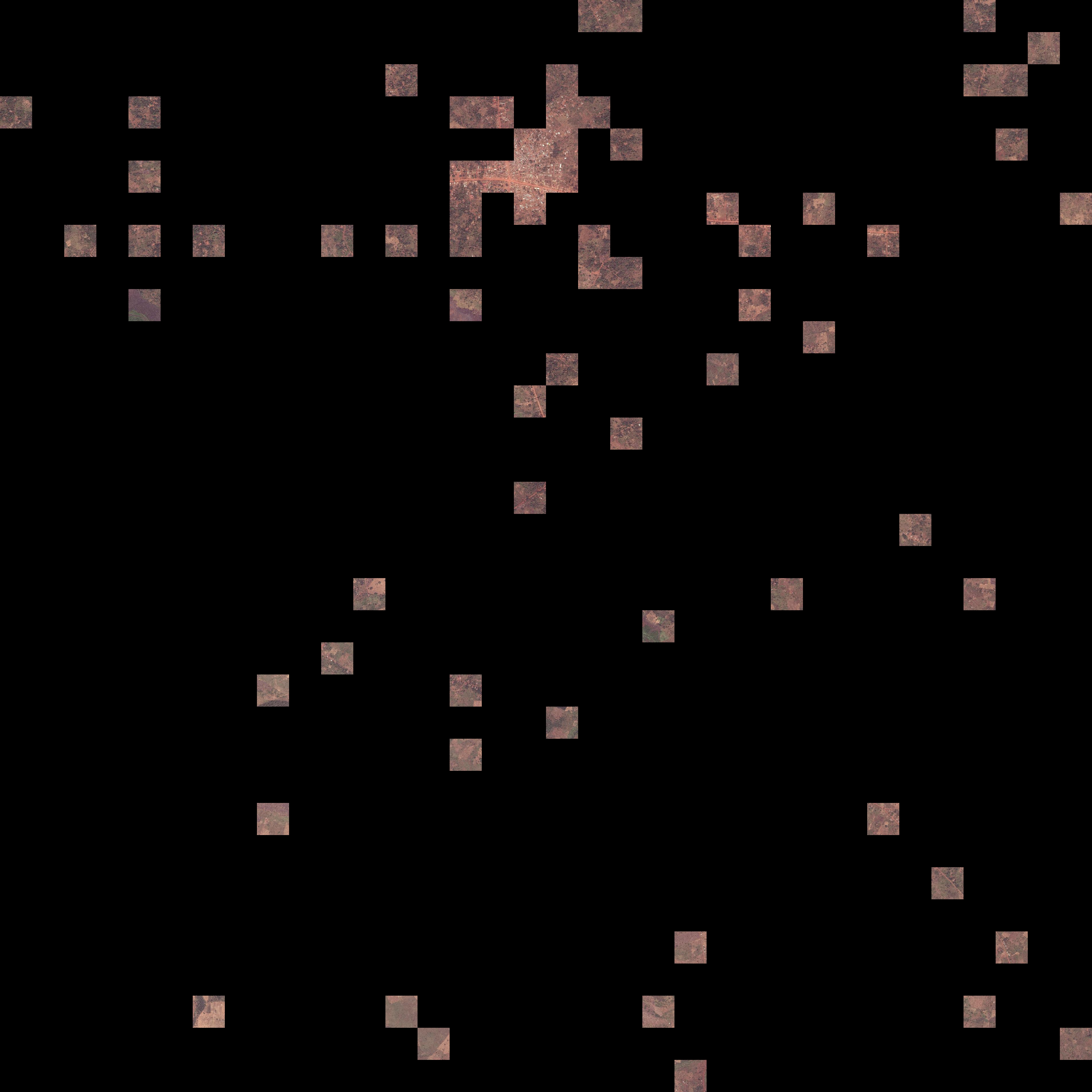}
    \caption{}
    \label{fig:wet_pred_sentinel1}
\end{subfigure}

\caption{(a) High-Resolution Satellite Imagery representing a cluster. (b) Sentinel-2 Imagery of the cluster from dry season. (c) Corresponding HR acquisitions when dry-season imagery is input to the Policy Network. (d) Sentinel-2 Imagery of the cluster from wet season. (e) Corresponding HR acquisitions when wet-season imagery is input to the Policy Network.}
\label{fig:dry_wet_comparison1}
\end{figure*}
\subsection{Testing the Policy Network on Poverty Prediction}
Previously, we trained and tested the policy network to quantify how well we approximate the true object counts. In this section, we train and test the policy network on Uganda dataset where we have only cluster-level poverty labels. 

\textbf{Poverty Estimation.}
Previous work \cite{ayush2020generating} exhaustively performed object detection on all the HR tiles representing a cluster $i$ to obtain $T$ $L$-dimensional vectors, $\mathbf{v}_i = \{\mathbf{v}_i^j\}_{j=1}^T$, which are then aggregated into a single $L$-dimensional categorical feature vector, $\mathbf{m}_i$, by summing over the tiles \emph{i.e.} $\mathbf{m}_i = \sum_{j=1}^{T} \mathbf{v}_i^j$. This was subsequently used in a regression model to predict poverty score for cluster $i$. Using our adaptive method, we obtain $\mathbf{\hat{m}}_{i} = \sum_{j=1}^{T} \mathbf{\hat{v}}_i^j$, which is an approximate classwise counts vector for cluster $i$. 
Following \cite{ayush2020generating}, we consider Gradient Boosting Decision Trees as the regression model to estimate the poverty index, $y_i$, given the cluster level categorical feature vector (classwise object counts), $\mathbf{m}_i$ or $\hat{\mathbf{m}}_i$.  

\begin{figure*}[!h]
    \centering
    \includegraphics[width=1.0\textwidth]{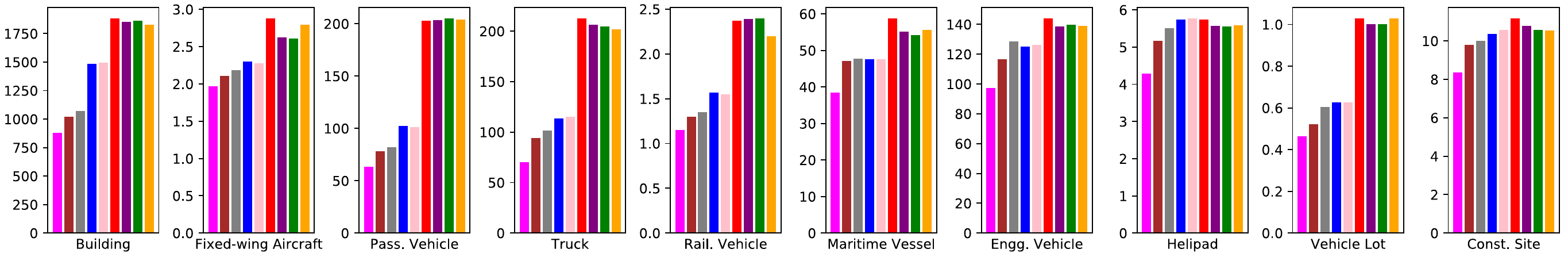}
    \caption{Number of objects missed on average across clusters for each class. Colored bars in each subplot from left-right are: \textcolor{magenta}{Ours (wet season)}, \textcolor{brown}{Ours (dry season)},
    \textcolor{gray}{Counts Pred.},
    \textcolor{blue}{Nightlight},
    \textcolor{pink}{Settlement},
    \textcolor{red}{Fixed-18}, \textcolor{purple}{Random-25},
    \textcolor{green}{Green Tiles},
    \textcolor{orange}{Stochastic-25}.
    }
    \label{fig:counts_diff}
\end{figure*}
\begin{figure*}[!h]
\centering
\begin{subfigure}[b]{0.22\textwidth}
    \includegraphics[width=\textwidth]{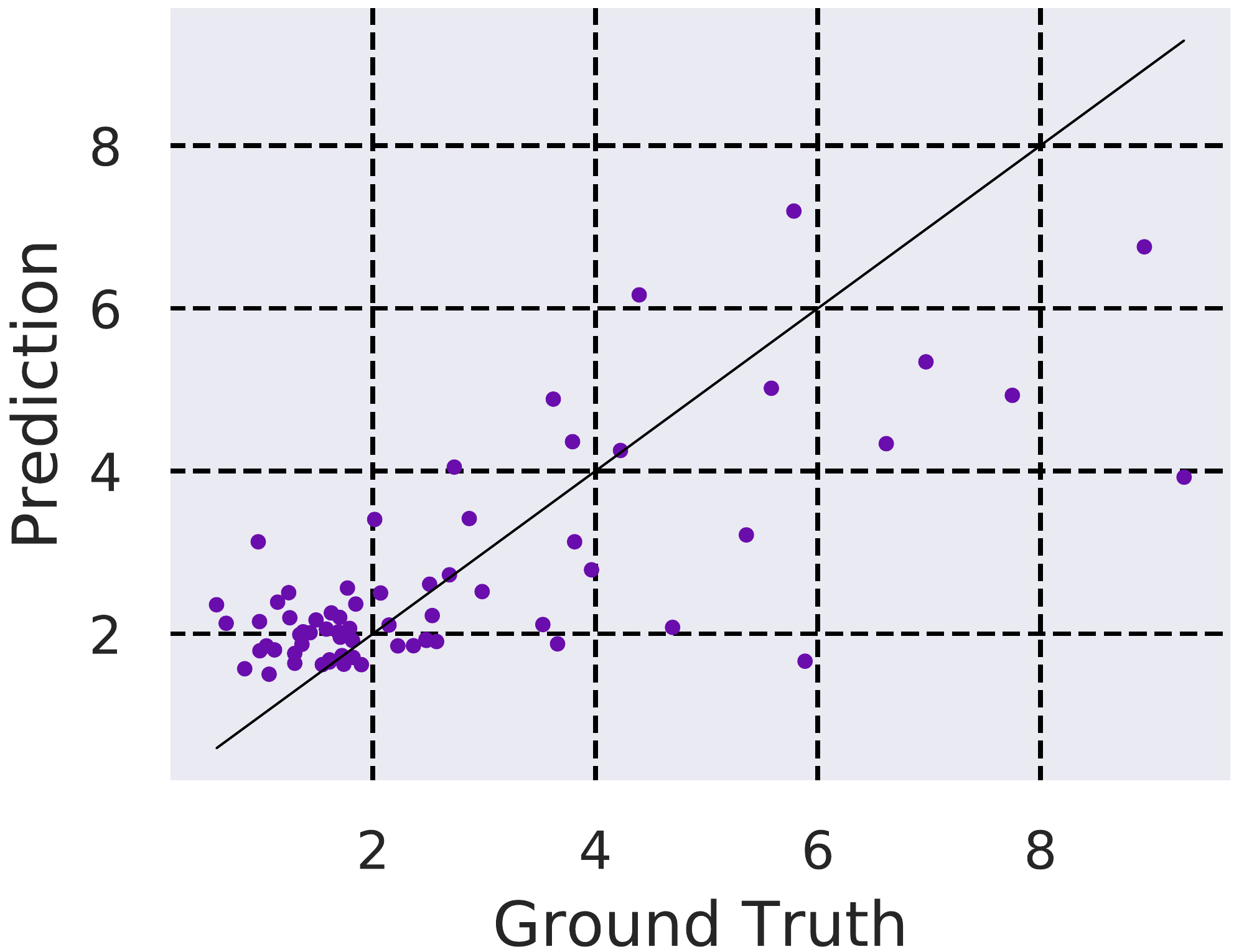}
    \caption{No Dropping}
    \label{fig:no_drop}
\end{subfigure}%
~
\begin{subfigure}[b]{0.22\textwidth}
    \includegraphics[width=\textwidth]{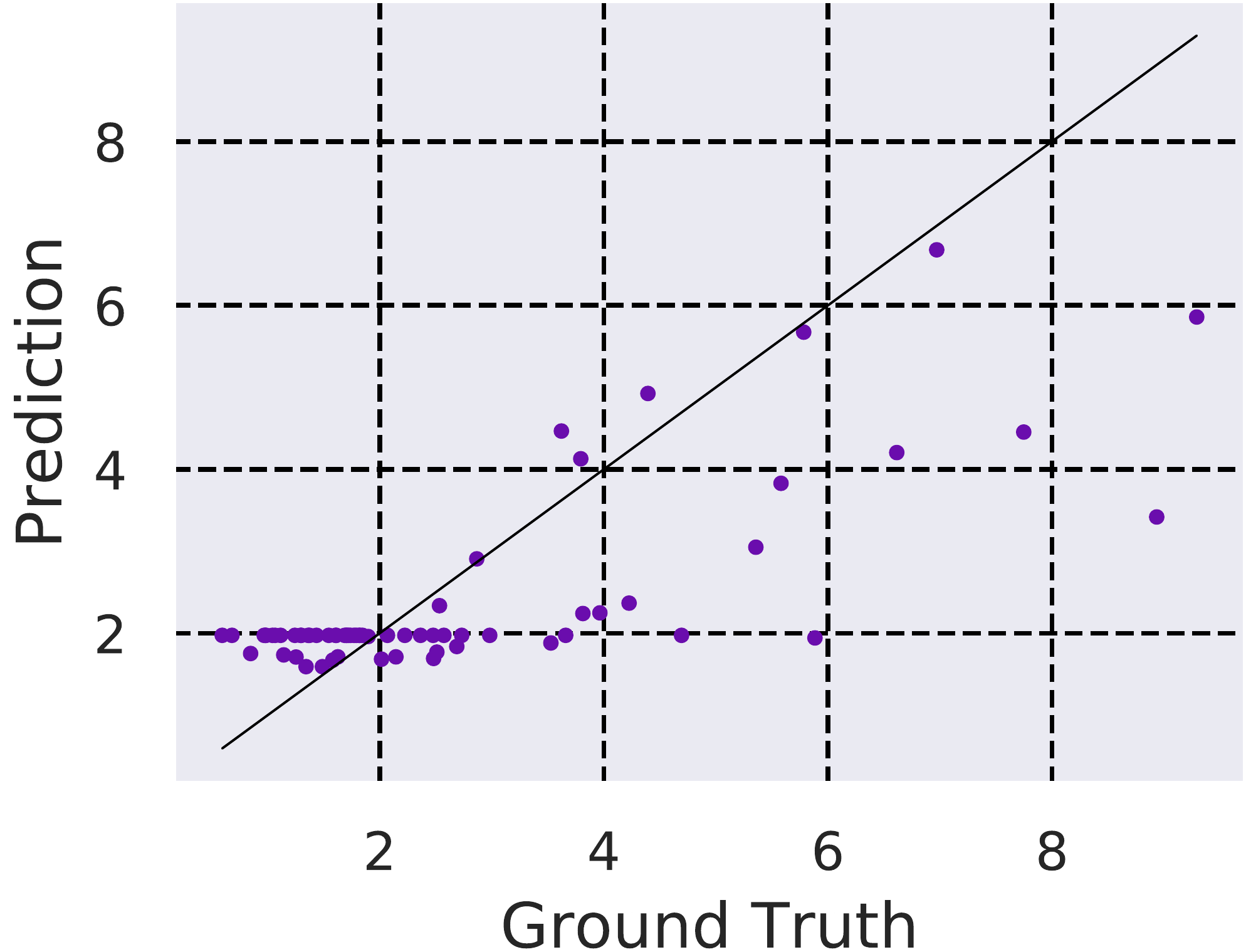}
    \caption{Nightlights}
    \label{fig:nightlight}
\end{subfigure}%
~
\begin{subfigure}[b]{0.22\textwidth}
    \includegraphics[width=\textwidth]{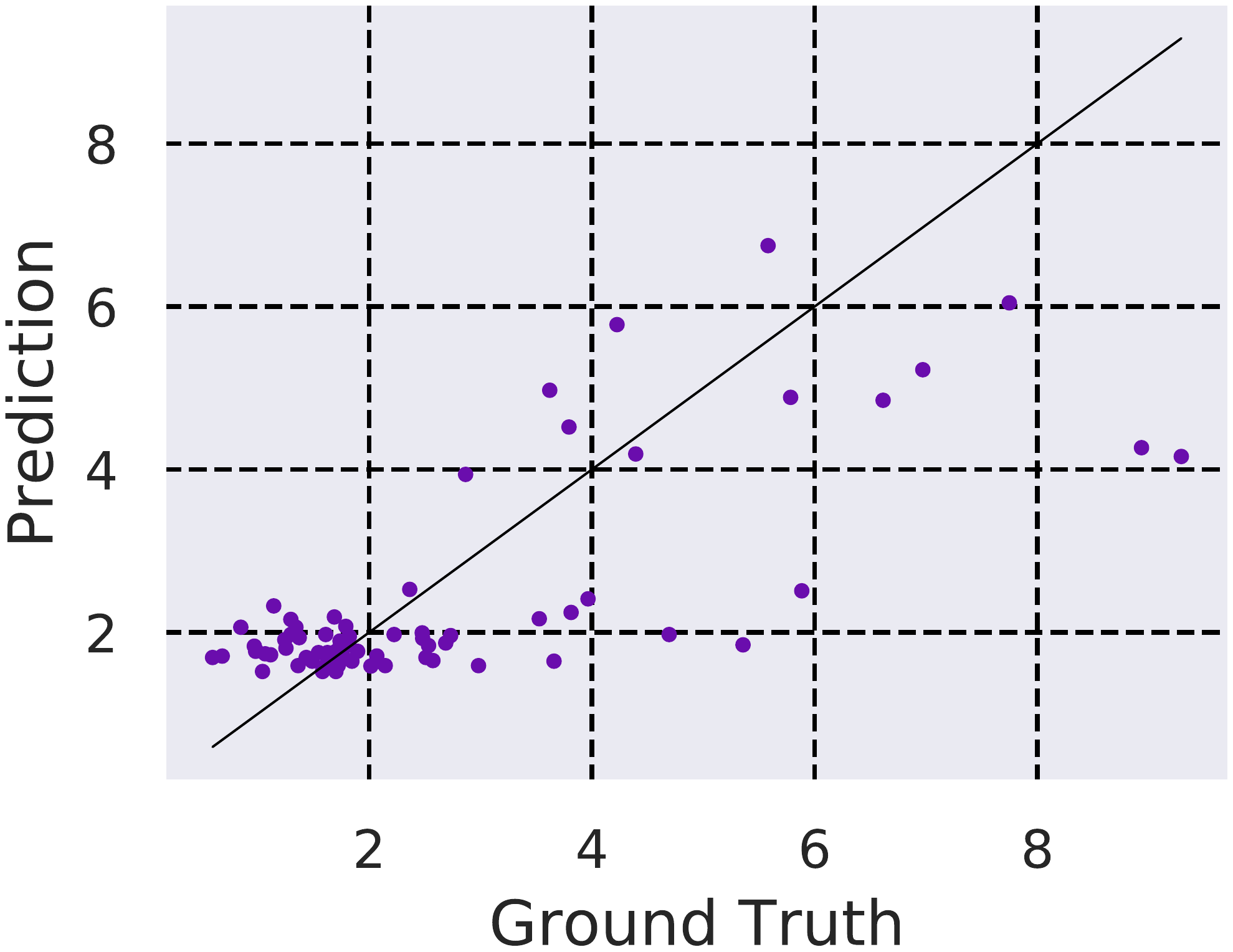}
    \caption{Ours (Dry Season)}
    \label{fig:dry_season}
\end{subfigure}%
~
\begin{subfigure}[b]{0.22\textwidth}
    \includegraphics[width=\textwidth]{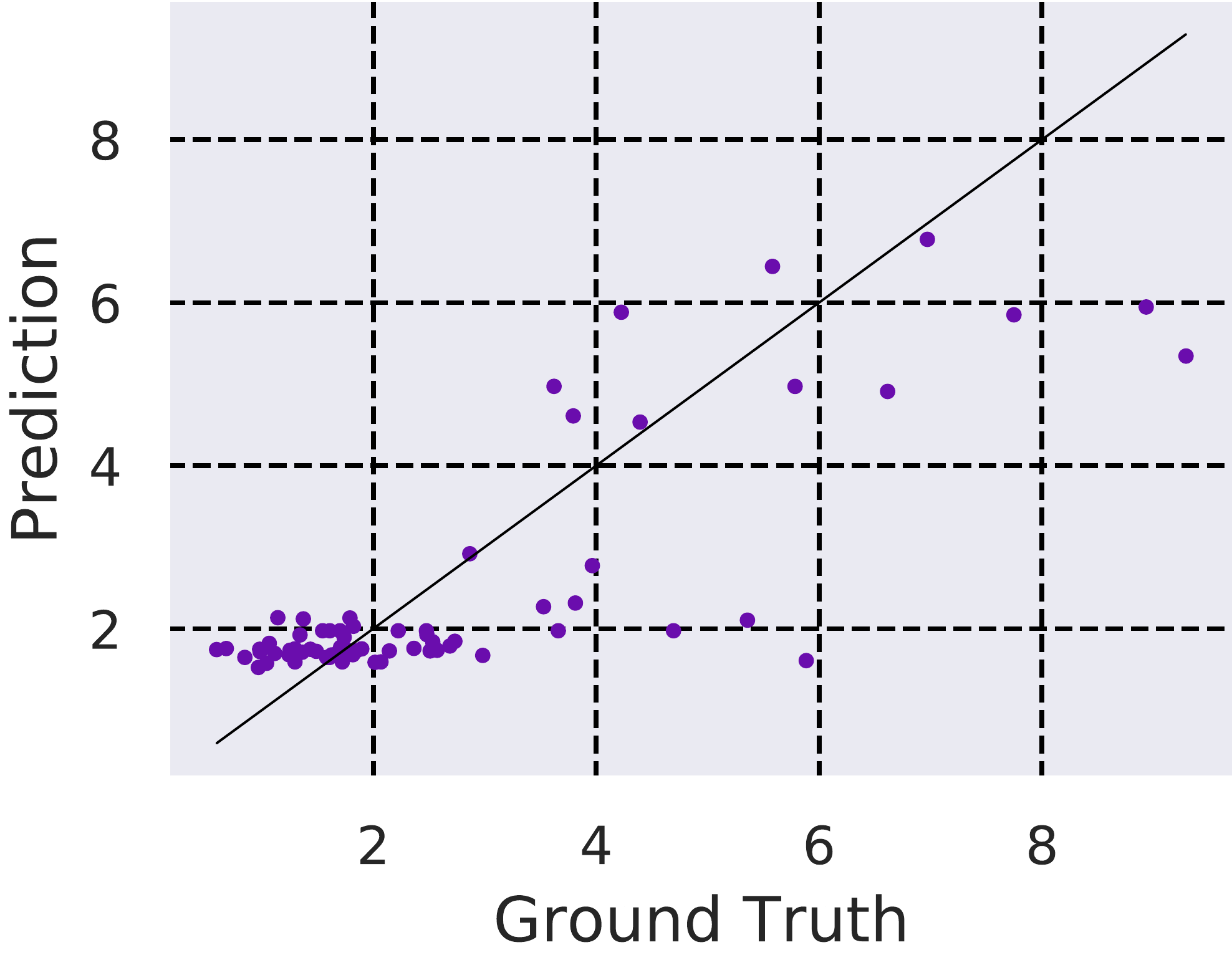}
    \caption{Ours (Wet Season)}
    \label{fig:wet_season}
\end{subfigure}
\caption{LSMS poverty score regression results of GBDT.}
\label{fig:predvsgt}
\end{figure*}
\begin{figure*}[!h]
\centering
\begin{subfigure}[b]{0.32\textwidth}
    \includegraphics[width=\textwidth]{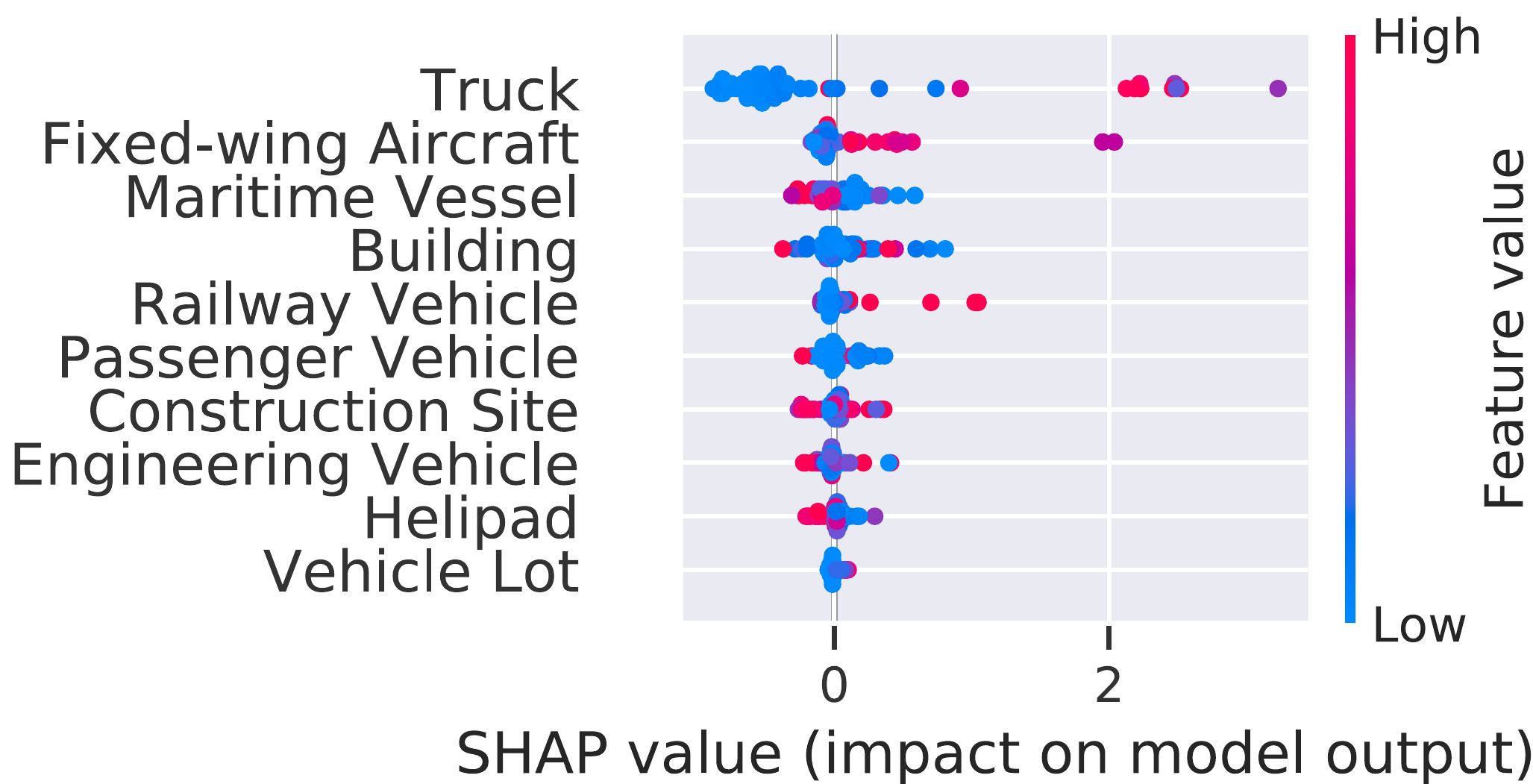}
    \caption{No Dropping}
    \label{fig:summary_all}
\end{subfigure}%
~
\begin{subfigure}[b]{0.32\textwidth}
    \includegraphics[width=\textwidth]{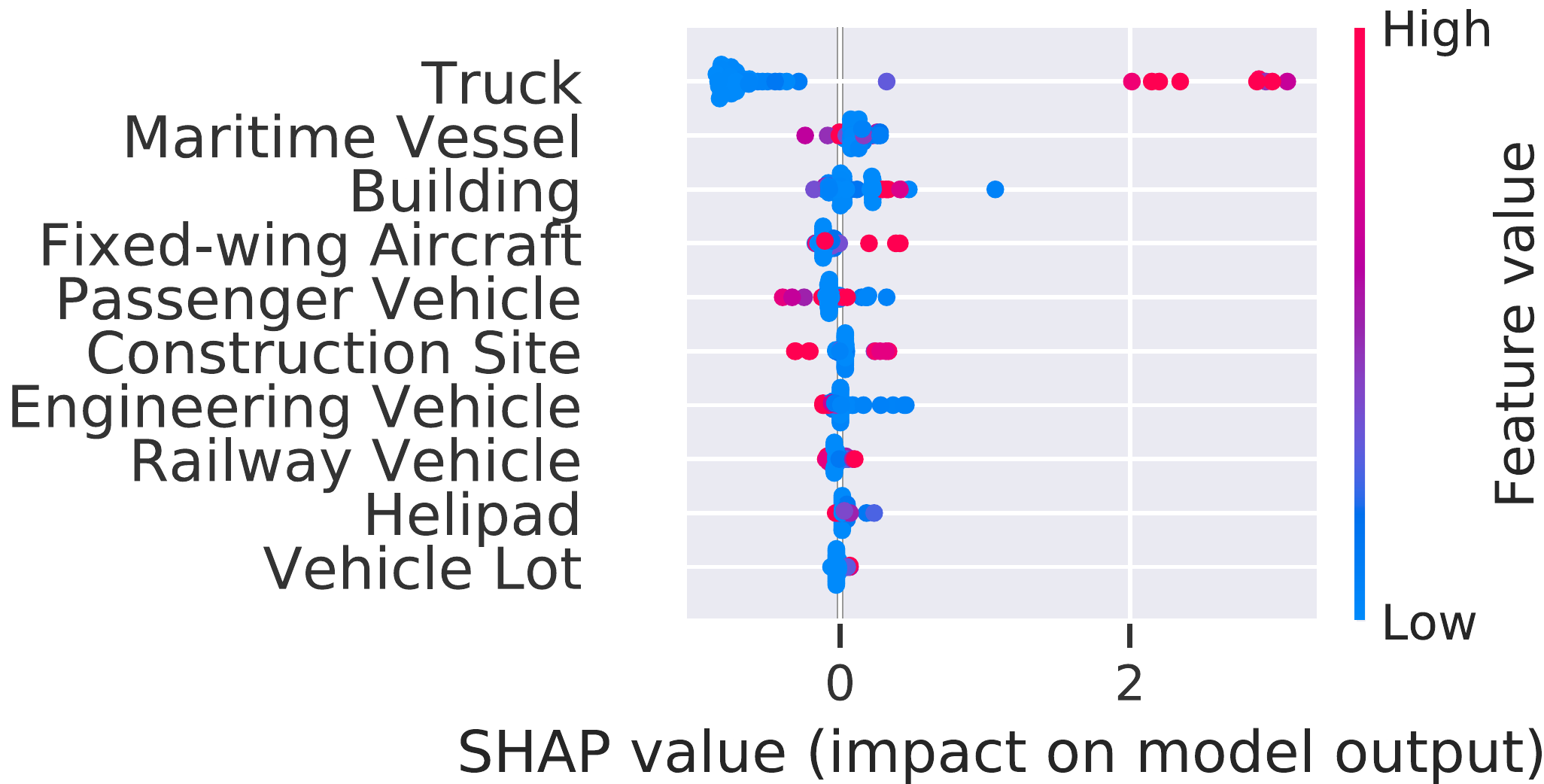}
    \caption{Ours (Dry Season)}
    \label{fig:summary_rl_wet}
\end{subfigure}
~
\begin{subfigure}[b]{0.32\textwidth}
    \includegraphics[width=\textwidth]{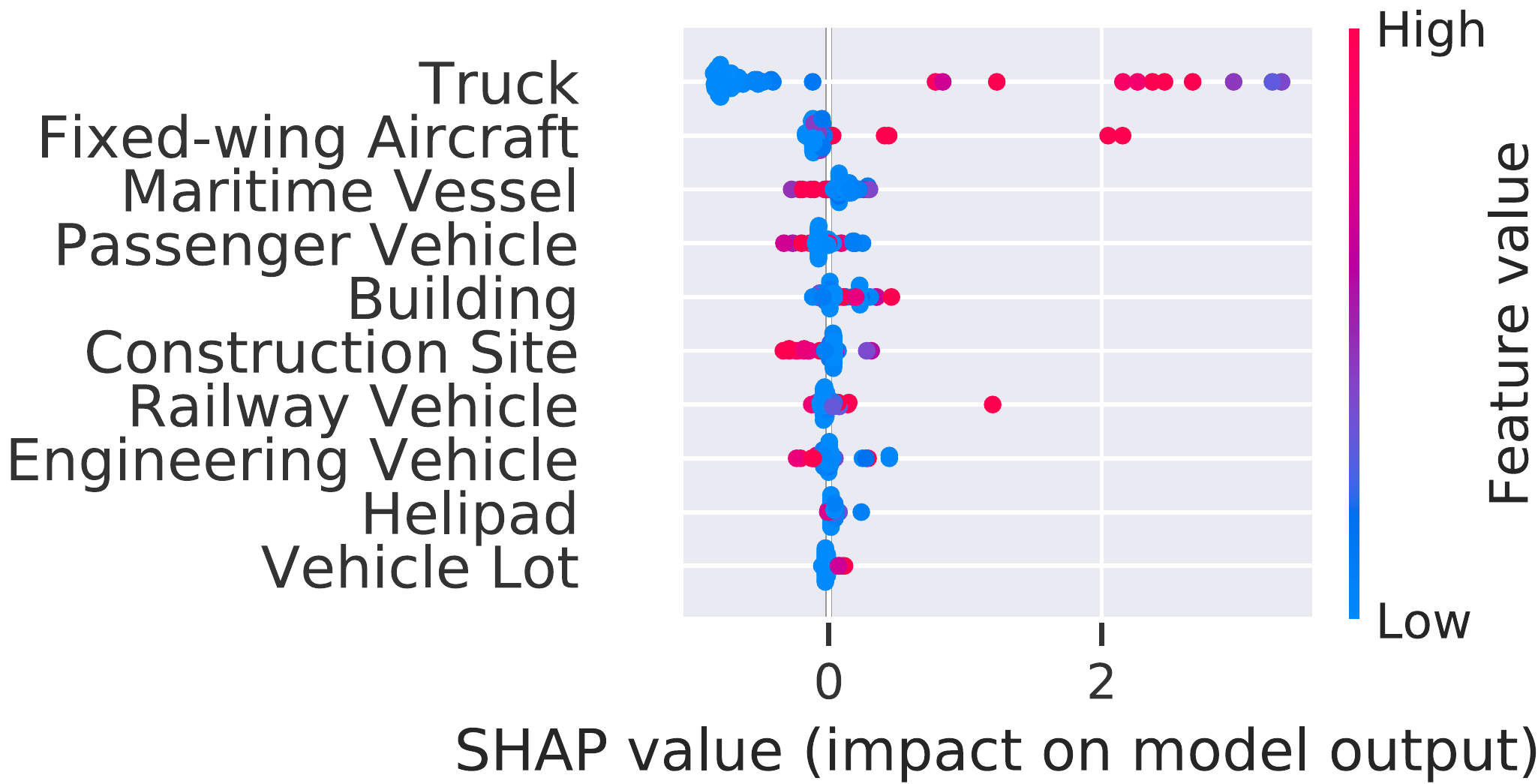}
    \caption{Ours (Wet Season)}
    \label{fig:summary_rl_dry}
\end{subfigure}
\caption{Summary of the effects of all features using SHAP, showing the distribution of the impacts each feature has on the model output. Color represents the feature value (red high, blue low).}
\label{fig:shap}
\end{figure*}
\textbf{Training and Evaluation.} We have N=320 clusters in the survey. We divide the dataset into a 80$\%$-20$\%$ train-test split. We train a GBDT model using object counts features ($\mathbf{m}_i$) based on all HR tiles of the clusters in the training set. 
We use the clusters in the training set to train the policy network for adaptive tile selection. The trained policy network is then used to acquire informative HR tiles for each test cluster \emph{i.e} for a test cluster $i$, the policy network selects HR tiles (subsequently used to obtain $\mathbf{\hat{m}}_i$) conditioned on low-resolution input representing the cluster. The obtained $\mathbf{\hat{m}}_i$ is then passed through the 
trained GBDT model to get the poverty score $y_i$. See appendix for more details. 
To evaluate the models, we use Pearson’s $r^2$ to quantify the model performance. Invariance under seperate changes in scale between two variables allows Pearson's $r^2$ to provide insights into the ability of the model at distinguishing poverty levels. We also report mean squared error (MSE) and Explained Variance~\cite{rosenthal2011statistics}. Explained variance measures the discrepancy between a model and actual data. Higher explained variance indicates a stronger strength of association thus meaning better predictions.

\textbf{Baselines and State-of-the-Art Models.}
We compare our method with the following: (a) \emph{No Patch Dropping}, where we simply use all the HR tiles in $\mathcal{H}_i$ to get the classwise object counts features (same as \cite{ayush2020generating}), (b) \emph{Fixed Policy-X} samples $X\%$ HR tiles from the center of a cluster, (c) \emph{Random Policy-X} samples $X\%$ HR tiles randomly from a cluster, (d) \emph{Stochastic Policy-X} samples $X\%$ HR tiles where the survival likelihood of a tile decays w.r.t the euclidean distance from the cluster center, (f) \emph{Green Tiles}, where we compute the average green channel value for a low-res tile and select bottom $K$ tiles for HR acquisition with least average green channel value, where $K$ is the number of tiles selected by the policy network for a particular cluster, (g) \emph{Counts Prediction}, where we train a CNN (Resnet-50 backbone) to regress object counts given low-res tile as input. We find that the object counts in a tile vary from 0-500. Instead of regressing directly on raw object counts, we create 100 bins such that a tile with counts between 5i-(5i+1) has label 5i+2.5 (e.g. a tile with counts 0-5 has label 2.5, 5-10 has label 7.5 and so on). We use this network to select top $K$ HR tiles based on predicted object counts, (h) \emph{Settlement Layer}, where we select HR tiles based on their population density. We used the HR settlement layer maps\footnote{https://research.fb.com/downloads/high-resolution-settlement-layer-hrsl/} and selected top $K$ tiles based on population density,
and (e) \emph{Nightlights}, where we use Nightlight Images ($48 \times 48$ px) representing the clusters in Uganda and sample only those HR tiles which have non-zero nighttime light intensities. 

Additionally, since Sentinel-2 imagery is freely available, we perform a comparative analysis of the effect of season on the ability of the policy network at approximating classwise object counts. We thus acquired two sets of low-resolution imagery, one from dry-season (Dec - Feb) in Uganda and other from wet season (March-May, Sept-Nov) corresponding to the survey year. Seasonality is likely highly relevant in our rural setting, where crops are grown during the wet season and much related market activity is highly seasonal. We hypothesize that greenery in low-resolution imagery during 
wet season will better indicate which patches might contain useful economic information.
\begin{figure}[!b]
    \centering
    \includegraphics[width=0.26\textwidth]{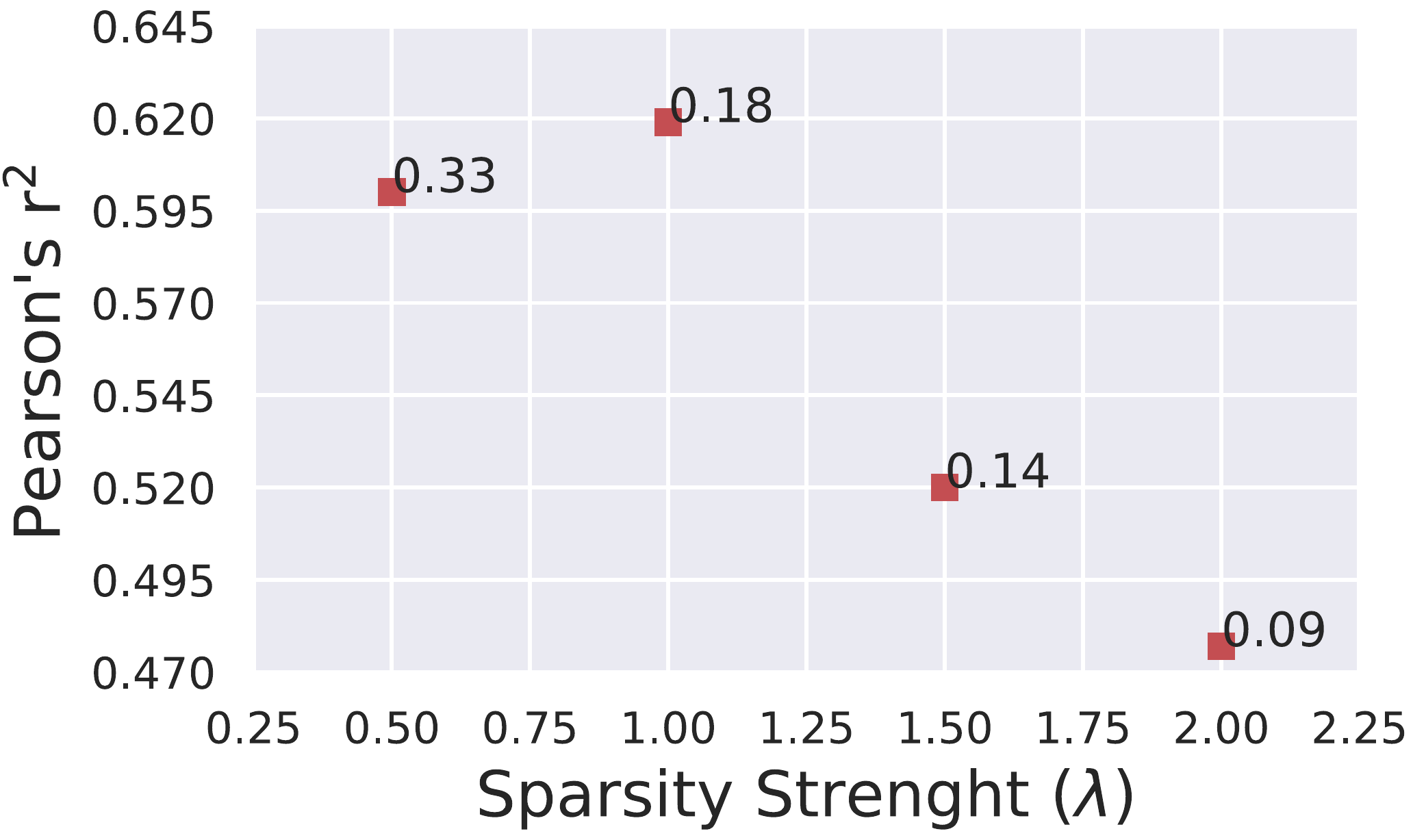}
    \caption{Trade-off between Pearson's $r^2$ and coefficent of image acquisition cost ($\lambda$). Text accompanying the points represents HR acquisition fraction.}
    \label{fig:trade_off}
\end{figure}

\textbf{Quantitative Analysis.} Fig. \ref{fig:counts_diff} compares the ability of various methods at approximating the classwise object counts. It shows the number of objects  missed on  an average across clusters for each parent class, where we can see that our method (using wet season imagery) can better approximate the ``true object counts'' (we use object detector predictions on all the HR tiles as a proxy for true values)
compared to baselines and our method (using dry season imagery).
Table~\ref{tab:comparison} shows the results of poverty prediction in Uganda. Our model (wet season) achieves \textbf{0.61} $r^2$ and substantially outperforms the published state-of-the-art results \cite{ayush2020generating} (\textbf{0.53} $r^2$) while using around \textbf{80\%} fewer HR images. 

It is interesting that we can outperform \emph{No Dropping} method when sampling only $20\%$ of HR tiles. Qualitatively, we observed that it is due to false positives proposed by the object detector on the tiles with no true objects of interest in it. Unfortunately, since we do not have ground truth bounding boxes for Uganda, we can not quantify it. However, our experiments on xView (Table~\ref{tab:xview_exps}) show that our approach achieves higher AP than the \emph{No Dropping} approach, suggesting our approach is able to remove false positives.

In comparison to the baselines relying on external data layers such as settlement and nightlights, our method achieves around 0.16 higher $r^2$. This is because such maps assume that objects are located in the tiles with large nightlight intensity or settlement index, however, some objects, i.e trucks, passenger vehicles etc., do not necessarily exist in these areas. Additionally, our approach outperforms the counts prediction model by 0.12 in $r^2$. This might be because the counts predictor is trained to directly regress very noisy object counts thus making it a difficult task. 

Next, a scatter plot of GBDT LSMS poverty score predictions v.s. ground truth is shown in Fig.~\ref{fig:predvsgt}. It can be seen that the GBDT model can maintain explainability of a large fraction of the variance based on object counts identified from the sampled HR tiles using our method, compared to \cite{ayush2020generating} that exhaustively uses all HR tiles.  

\textbf{Performance$/$Sampling Trade-off.} We analyze the trade-off between accuracy (regression performance) and HR sampling rate controlled by the hyperparameter $\lambda$ in the reward Eq.~\ref{eq:tradeoff}. We intentionally change $\lambda$ to quantify the effect on the policy network. As seen in Fig.~\ref{fig:trade_off}, the policy network samples less HR tiles (a 0.09 fraction) when we increase $\lambda$ to 2.0 and the $r^2$ goes down to 0.48. On the other hand, when we set $\lambda$ to 1.0, we get optimal results in terms of $r^{2}$, while acquiring only a 0.18 fraction of HR imagery. 

\textbf{Cost saving}. Current pricing for high-resolution (30cm) RGB imagery is 10-20\$ per km$^2$. Given that Uganda is 240k km$^2$ in land area, creating a poverty map using our method would save roughly \$2.9 million if imagery costs \$15 per km$^2$. 
This represents a potentially large cost saving if our approach is scaled at country or continent scale. 

\textbf{Analysis based on Season.}  Presence of greenery during wet season allows the policy network to better identify the informative regions containing objects, compared to when trained with dry season Sentinel-2 imagery as input. Fig.~\ref{fig:dry_wet_comparison1} presents an example cluster, where it is seen that training the policy network using wet season imagery better assists the network at sampling informative tiles (see Appendix).



\textbf{Impact on Interpretability.} An important contribution of \cite{ayush2020generating} was to introduce model interpretability allowing successful application of such methods in many policy domains.
They use Tree SHAP (Tree SHapley Additive exPlanations) \cite{NIPS2017_7062}, a game theoretic approach to explain the output of tree-based models, to explain the effect of individual features on poverty predictions. Here, we show that in addition to closely approximating the classwise object counts, our method retains the same findings for interpretability as that of \cite{ayush2020generating}. Fig. \ref{fig:shap} shows the plots of SHAP values of every feature for every cluster for three different methods. The features are sorted by the sum of SHAP value magnitudes over all samples. It can be seen that our method still maintains that \emph{\#Trucks} tends to have a higher impact on the model's output. We also observe that ordering of features in terms of SHAP values is fairly similar between the \emph{No Dropping} approach ~\cite{ayush2020generating} and our method.

\section{Conclusion}
In this study, we increase the efficiency of recent methods of predicting consumption expenditure using object counts from high-resolution satellite images. To achieve this, we proposed a novel reinforcement learning setup to conditionally acquire high-resolution tiles. 
We designed a cost-aware reward function to reflect real-world constraints -- i.e. budget and GPU availability -- and then trained a policy network to approximate object counts in a given location as closely as possible given these constraints.
We show that our approach reduces the number of high-resolution images needed by 80\% while improving downstream poverty estimation performance relative to multiple other approaches, including a method that exhaustively uses all high-resolution images from a location. Future work includes application of our adaptive method to other sustainability-related computer vision tasks using high-resolution images at large scale.

{\small
\bibliography{main.bib}
}


\end{document}


\maketitle
\appendix

\section{Pseudocode}
\label{sec:pseudocode}
\begin{algorithm}[H]
\hline
\vspace{0.3em}
\SetAlgoLined
\KwIn{($\mathcal{L}_{i}$, $\mathcal{H}_{i}$)\quad 
$i=\{1, 2, ..., N\}$ \\}
\For{$j\gets 1\:to\:T$}{
    $s_{i}^{j} \gets f_{p}(l_{i}^{j};\theta_{p})$ \\
    $s_{i}^{j} \gets \alpha + (1-s_{i}^{j})(1-\alpha)$ \\
    $\mathbf{a}_{i}^{j} \sim \pi(\mathbf{a}_{i}^{j}|s_{i}^{j})$ \\
    \BlankLine
    \For{$k\gets 1\:to\:S$}{
    $\mathbf{\hat{v}}_{i}^{j,k} = f_{d}(h_{i}^{j,k})\bigodot\:\mathbf{a}_{i}^{j,k}$ \\
    }
    $\mathbf{\hat{v}}_i^j = \sum_{k=1}^S \mathbf{\hat{v}}_i^{j,k}$
    \BlankLine
    \textbf{Evaluate Reward} $R(\mathbf{a}_{i}^{j}, \mathbf{\hat{v}}_{i}^{j}, \mathbf{v}_{i}^{j})$ \\
    \BlankLine
    $\theta_{p} \gets \theta_{p} + \nabla{\theta_{p}}$
}
\hline
\vspace{0.2em}
\caption{Pseudo-code for the Proposed Adaptive Algorithm. $T$ and $S$ represent the number of tiles and subtiles.}
\label{alg:tileDrop_Algorithm}
\end{algorithm}

\section{Implementation Details}
\paragraph{Policy Network.} To parameterize the policy network, we use ResNet-32 \cite{he2016deep} pretrained on the ImageNet dataset~\cite{russakovsky2015ilsvrc}. We train the policy network using 2 NVIDIA 1080ti GPUs and use the following hyperparameters: learning rate = 1e-4, \#epochs = 300, batch size = 289, coefficient of image acquisition cost \emph{i.e.} $\lambda$=1.0, temperature scaling $\alpha$ is gradually increased from 0.6 to 0.95.

\paragraph{Object Detector.} We use YOLOv3 architecture~\cite{redmon2018yolov3} as the object detector, chosen for its reasonable trade off between accuracy on small objects and run-time performance. The backbone network, DarkNet-53, is pre-trained on ImageNet. Following \cite{ayush2020generating}, we perform transfer learning by training the detector on xView dataset and running it on the Uganda HR patches. The object detections are obtained at 0.6 confidence threshold following~\cite{ayush2020generating}.

\section{Additional Qualitative Results}
\label{sec:visualizations}
\begin{figure*}[!h]
\centering
\begin{subfigure}[b]{0.40\textwidth}
    \includegraphics[width=\textwidth]{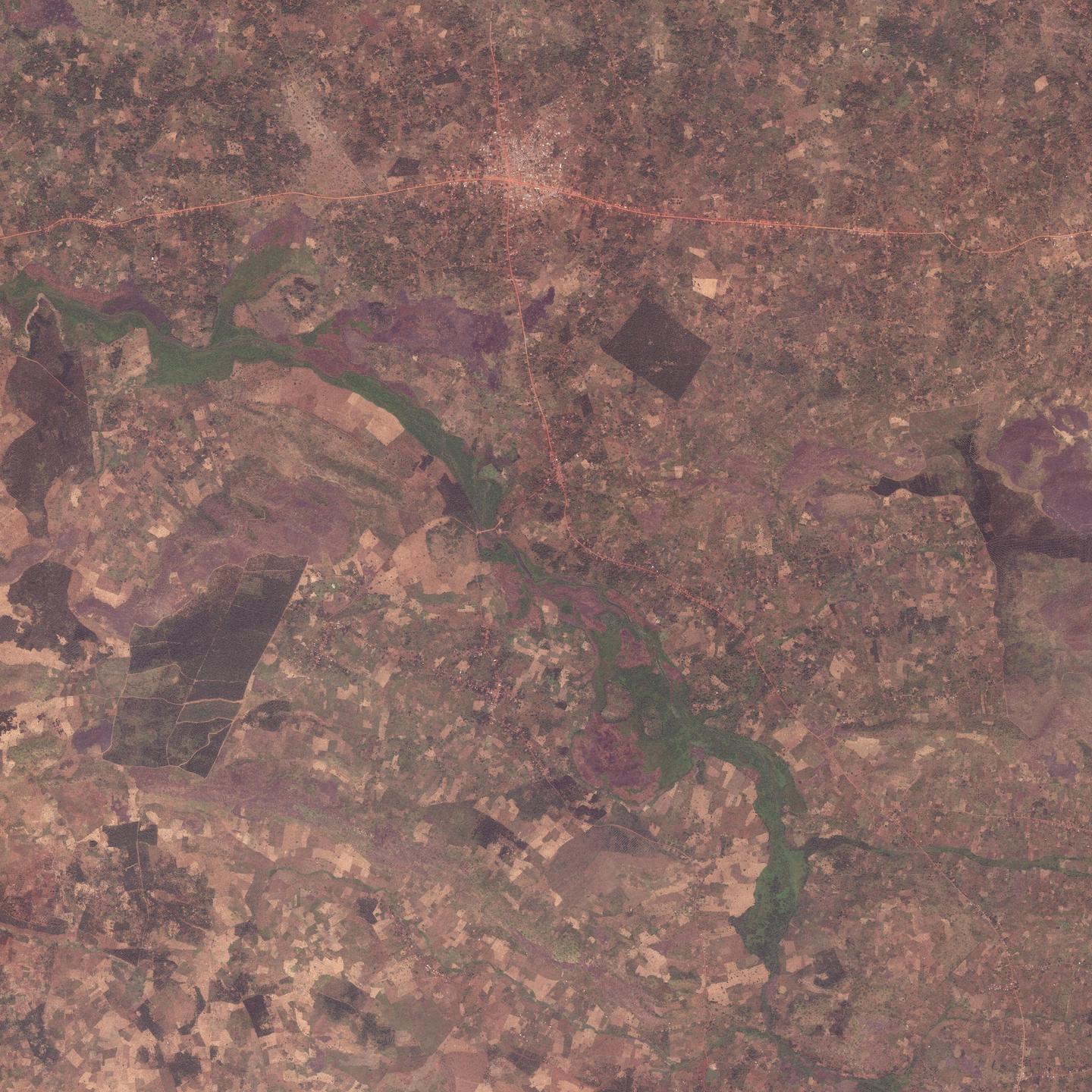}
    \caption{High-Resolution Satellite Imagery (downsampled for visualization) corresponding to a cluster.}
    \label{fig:original_dg1}
\end{subfigure}%
\\
\begin{subfigure}[b]{0.35\textwidth}
    \includegraphics[width=\textwidth]{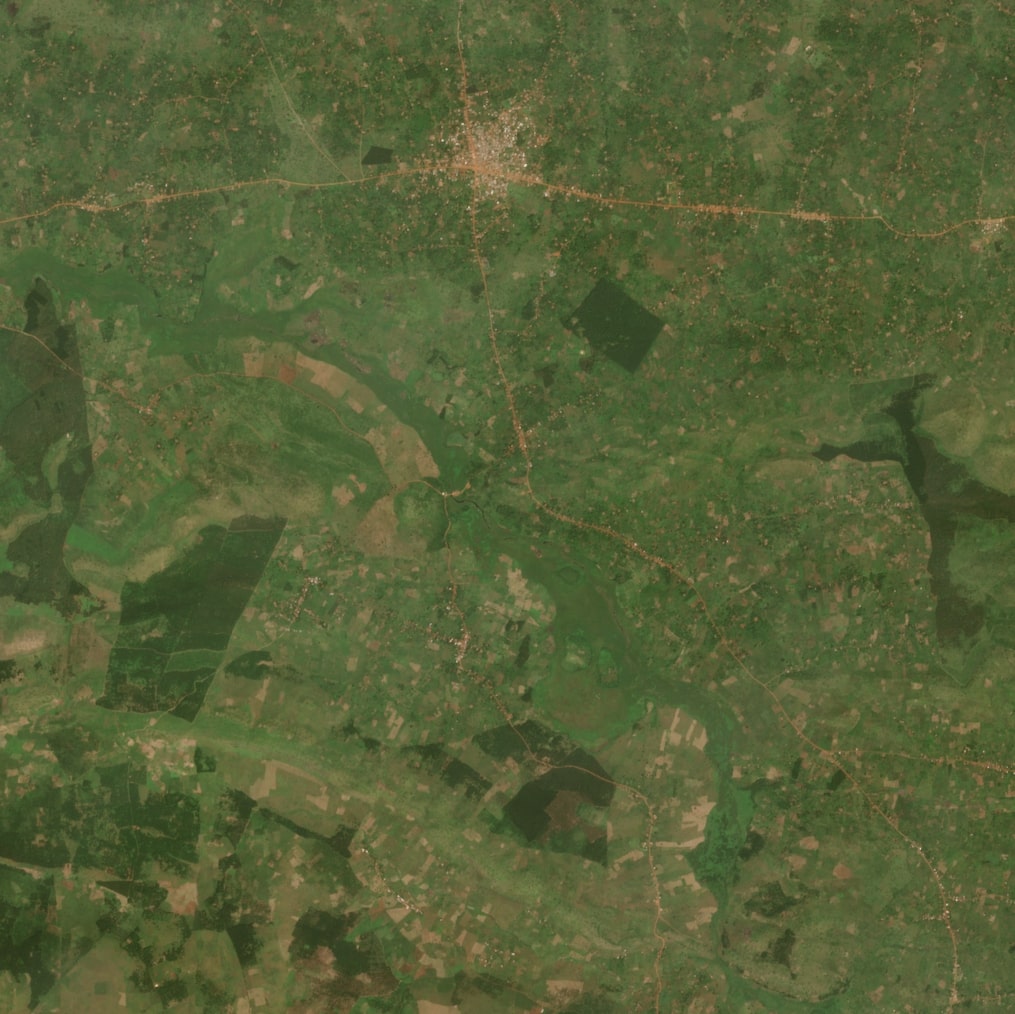}
    \caption{Low Resolution Sentinel-2 Imagery for the cluster from dry season.}
    \label{fig:dry_sentinel1}
\end{subfigure}%
~
\begin{subfigure}[b]{0.35\textwidth}
    \includegraphics[width=\textwidth]{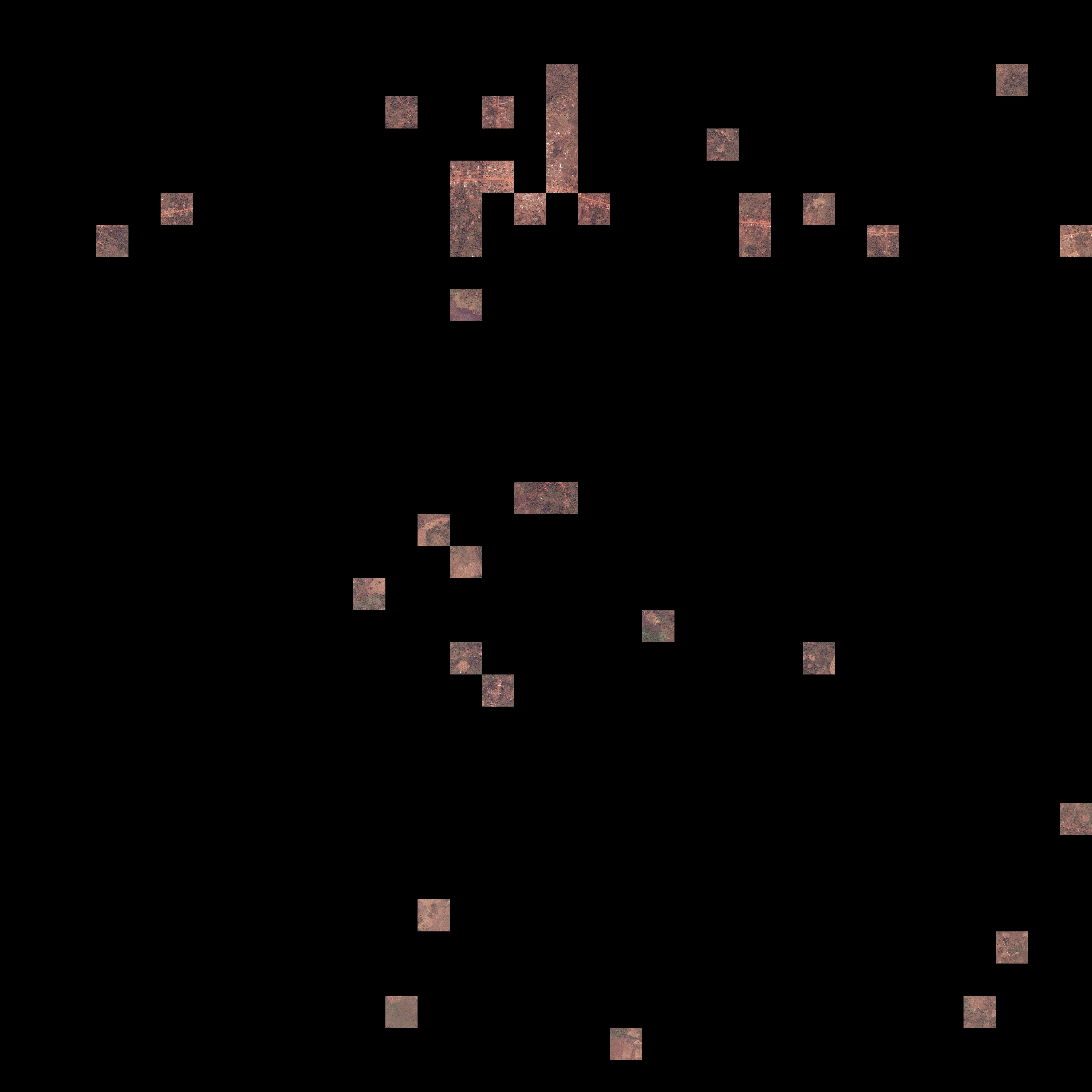}
    \caption{Corresponding HR acquisitions when dry-season imagery is input to the Policy Network.}
    \label{fig:dry_pred_sentinel1}
\end{subfigure}
~
\begin{subfigure}[b]{0.35\textwidth}
    \includegraphics[width=\textwidth]{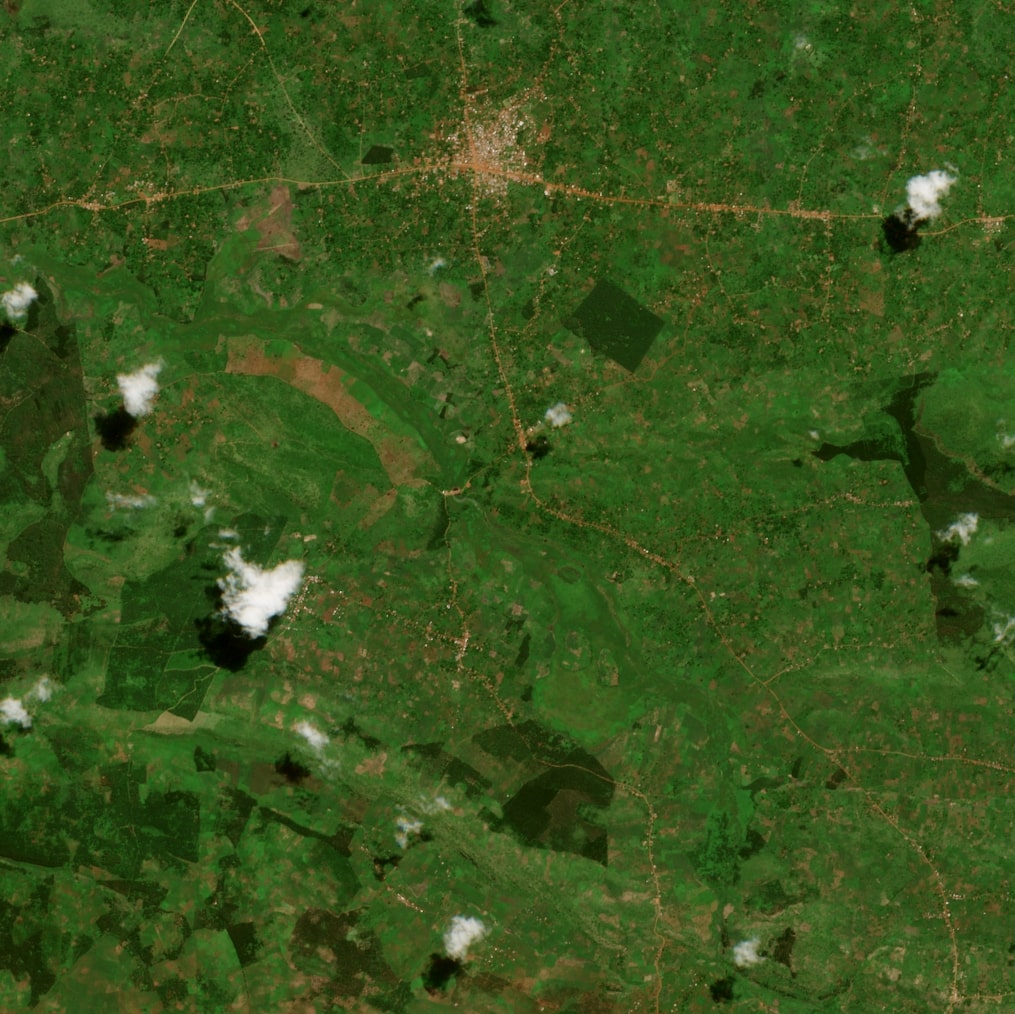}
    \caption{Low Resolution Sentinel-2 Imagery for the cluster from wet season.}
    \label{fig:wet_sentinel1}
\end{subfigure}
~
\begin{subfigure}[b]{0.35\textwidth}
    \includegraphics[width=\textwidth]{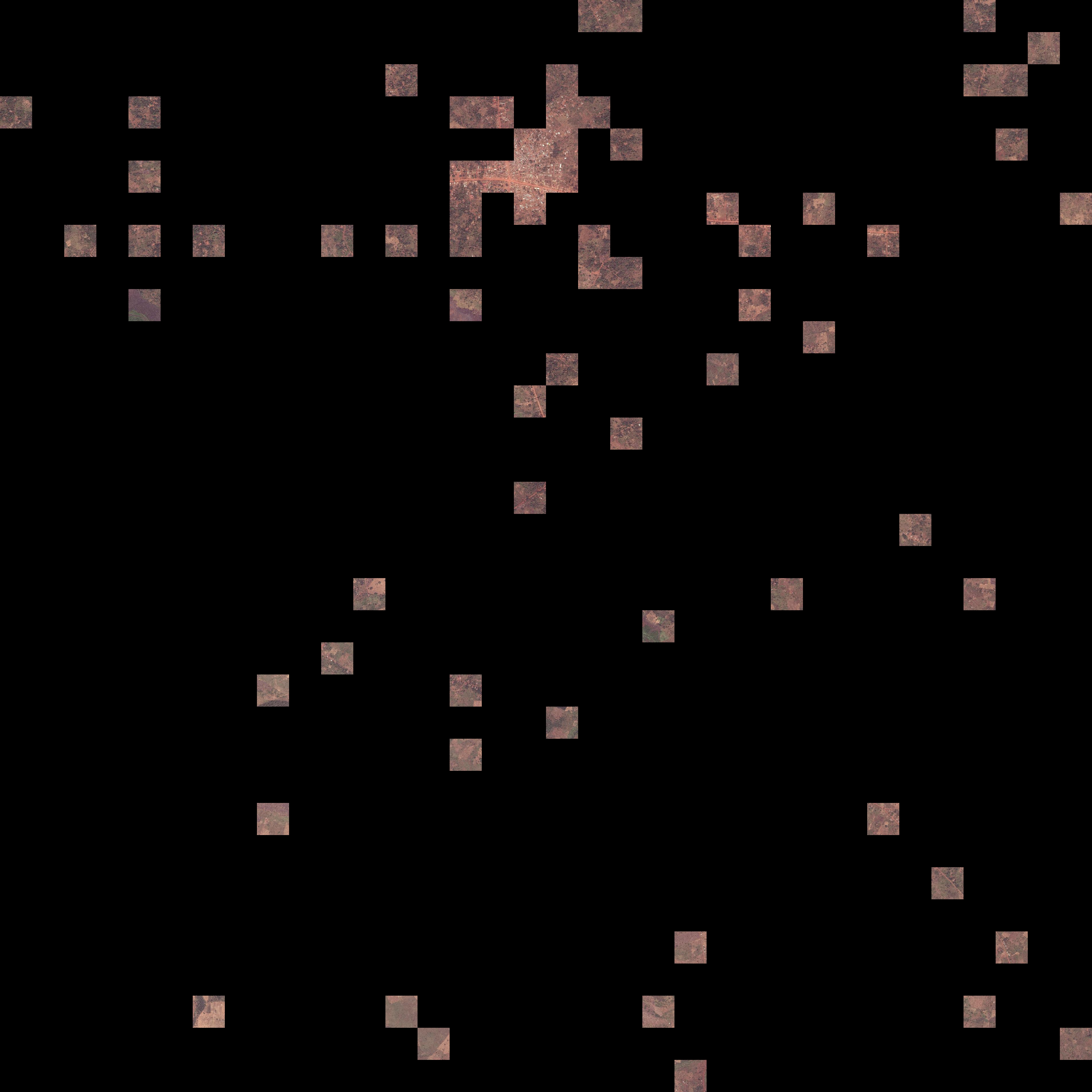}
    \caption{Corresponding HR acquisitions when wet-season imagery is input to the Policy Network.}
    \label{fig:wet_pred_sentinel1}
\end{subfigure}

\caption{Comparison between sampling ability of the policy network when trained with low-resolution imagery from two different seasons.}
\label{fig:dry_wet_comparison1}
\end{figure*}

\begin{figure*}[!h]
\centering
\begin{subfigure}[b]{0.40\textwidth}
    \includegraphics[width=\textwidth]{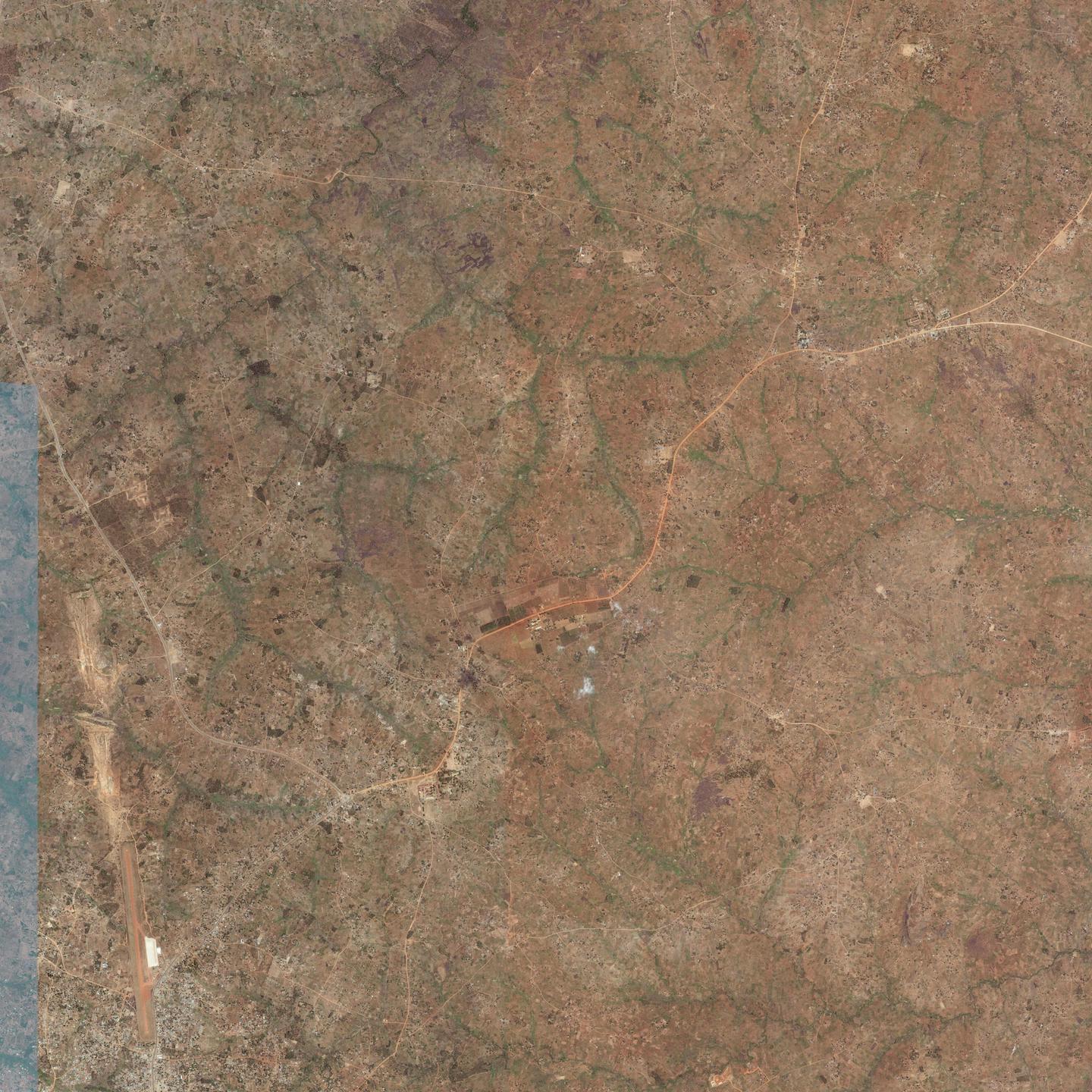}
    \caption{High-Resolution Satellite Imagery (downsampled for visualization) corresponding to a cluster.}
    \label{fig:original_dg2}
\end{subfigure}%
\\
\begin{subfigure}[b]{0.35\textwidth}
    \includegraphics[width=\textwidth]{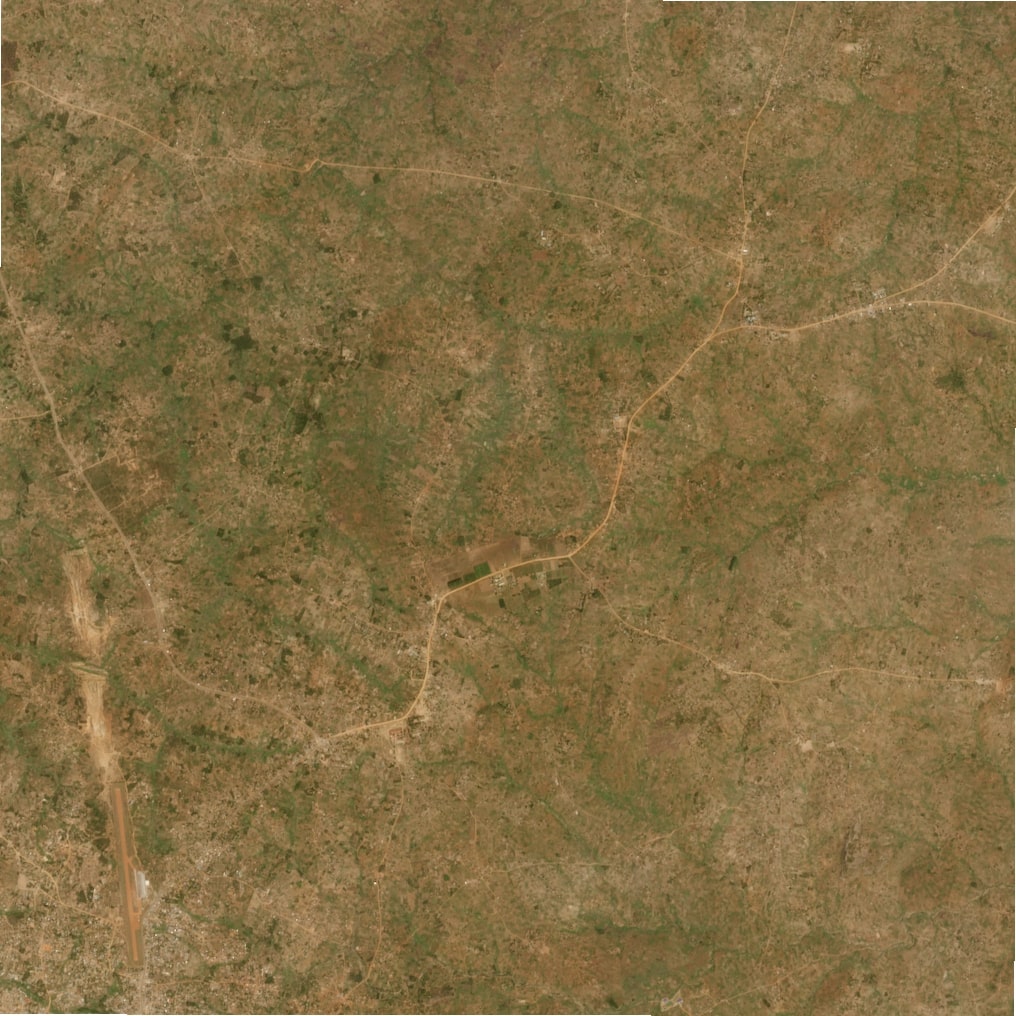}
    \caption{Low Resolution Sentinel-2 Imagery for the cluster from dry season.}
    \label{fig:dry_sentinel2}
\end{subfigure}%
~
\begin{subfigure}[b]{0.35\textwidth}
    \includegraphics[width=\textwidth]{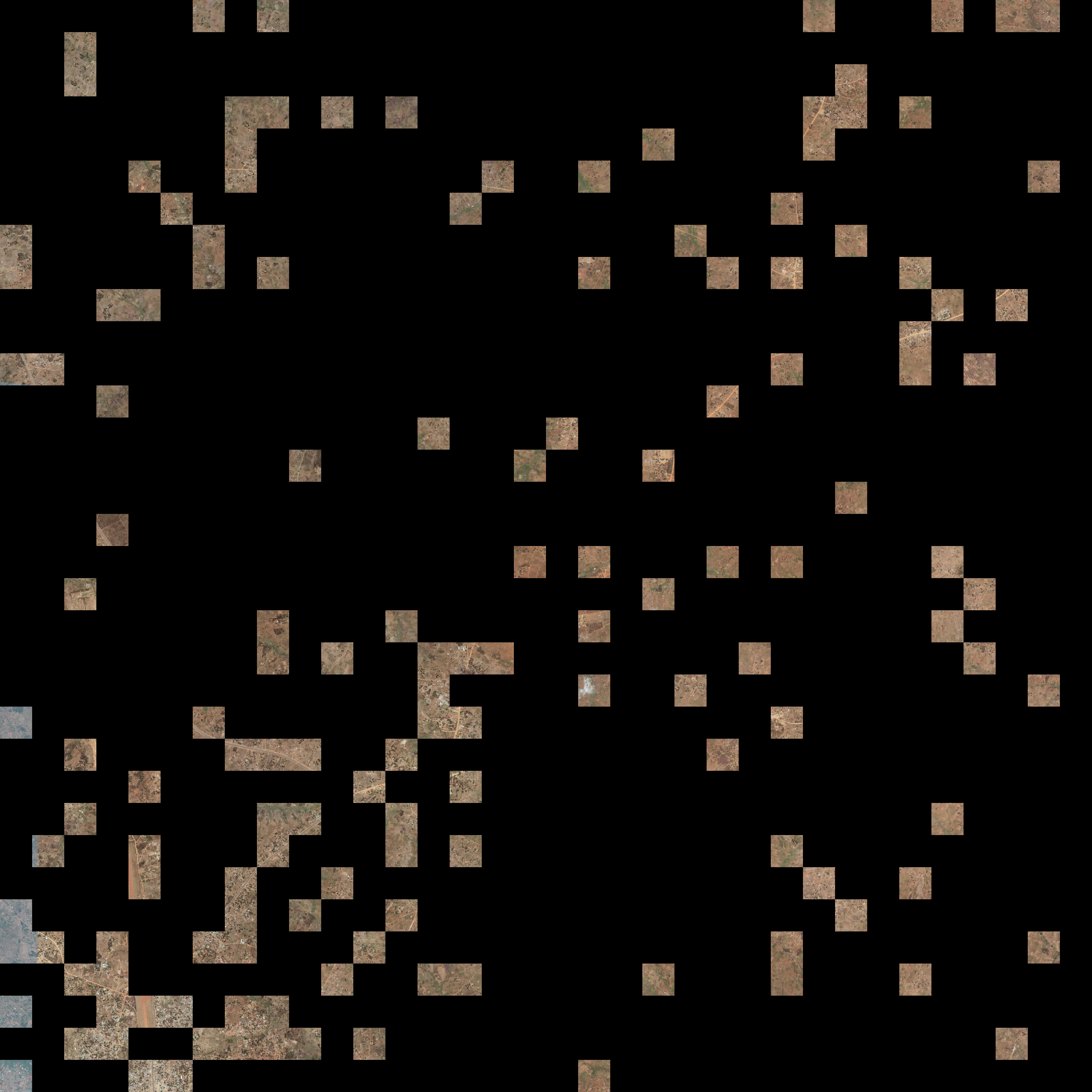}
    \caption{Corresponding HR acquisitions when dry-season imagery is input to the Policy Network.}
    \label{fig:dry_pred_sentinel2}
\end{subfigure}
~
\begin{subfigure}[b]{0.35\textwidth}
    \includegraphics[width=\textwidth]{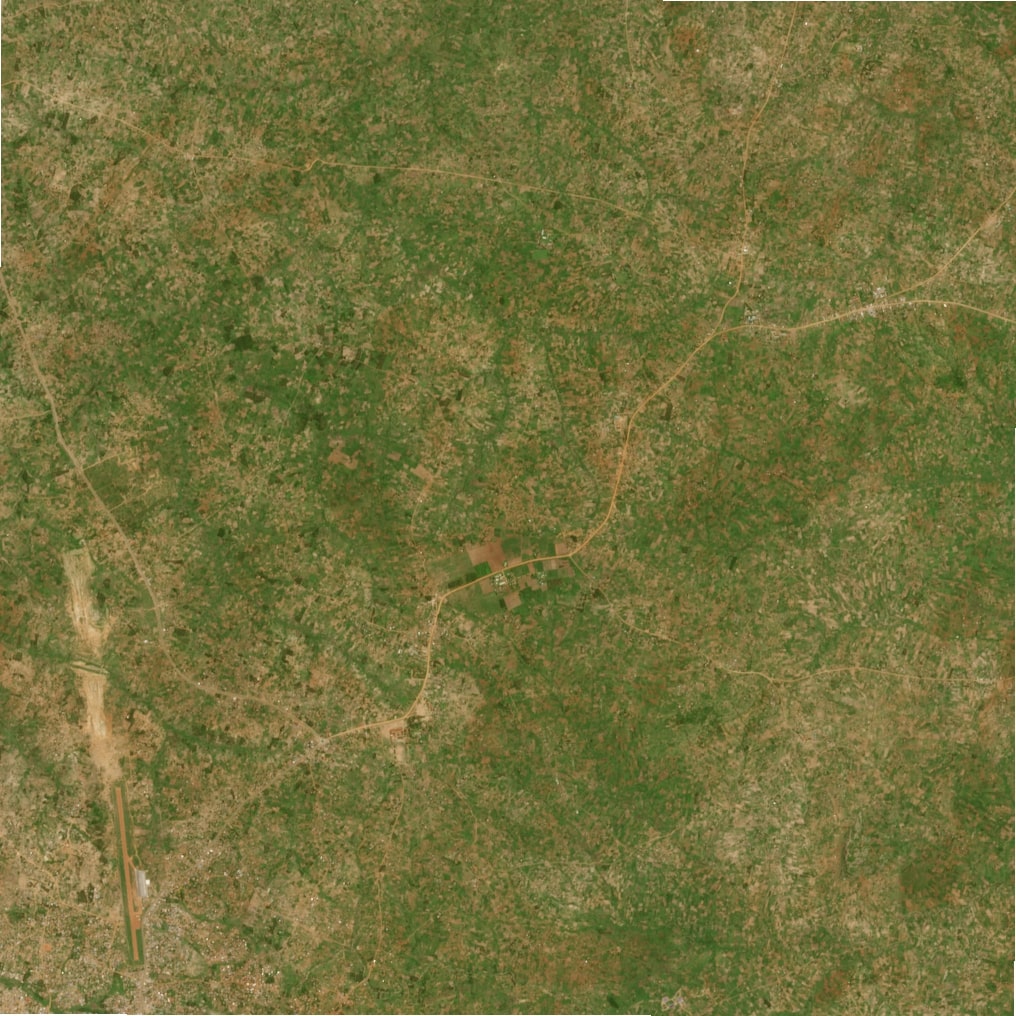}
    \caption{Low Resolution Sentinel-2 Imagery for the cluster from wet season.}
    \label{fig:wet_sentinel2}
\end{subfigure}
~
\begin{subfigure}[b]{0.35\textwidth}
    \includegraphics[width=\textwidth]{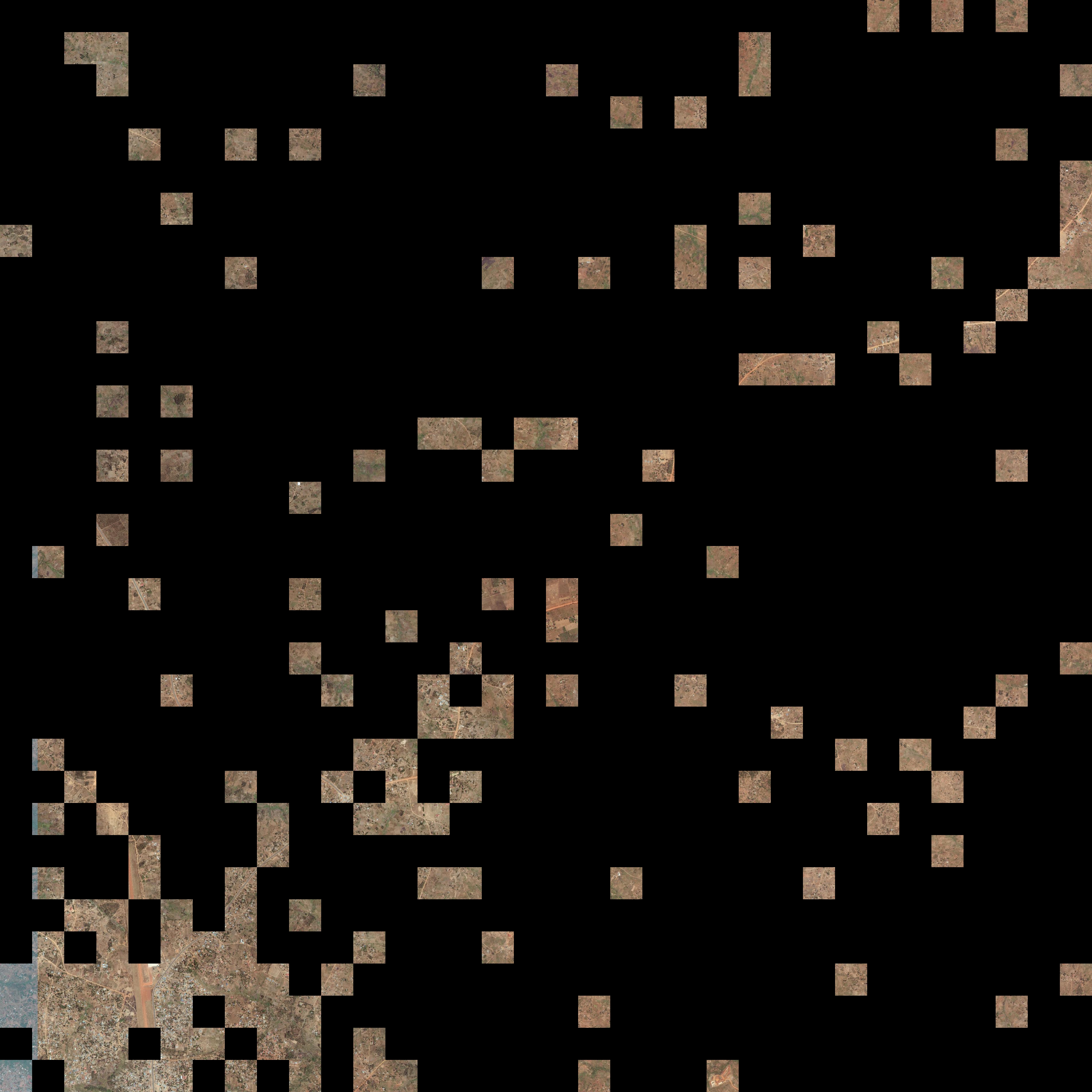}
    \caption{Corresponding HR acquisitions when wet-season imagery is input to the Policy Network.}
    \label{fig:wet_pred_sentinel2}
\end{subfigure}

\caption{Comparison between sampling ability of the policy network when trained with low-resolution imagery from two different seasons.}
\label{fig:dry_wet_comparison2}
\end{figure*}

In this section, we present additional qualitative results. Figures \ref{fig:dry_wet_comparison1} and \ref{fig:dry_wet_comparison2} emphasize that the presence of greenery during wet season allows the policy network to better identify the informative regions containing objects (better sampling of regions containing buildings, trucks, vehicles, etc.), compared to when dry season Sentinel-2 imagery is used as input to the network. Figures \ref{fig:sample_visualizations1}, \ref{fig:sample_visualizations2}, and \ref{fig:sample_visualizations3} show additional results when wet season imagery is used as input to the policy network.

\section{Code}
The code used for our experiments is given \href{https://anonymous.4open.science/r/b290d0ab-8a54-4457-8f19-cef497b7e2ab/}{here}.

\begin{figure*}[!h]
\centering
\begin{subfigure}[b]{0.55\textwidth}
    \includegraphics[width=\textwidth]{images/samples/10210013_all.jpeg}
    \caption{High-Resolution Satellite Imagery (from Digital Globe Systems) corresponding to a cluster.}
    \label{fig:dgs1}
\end{subfigure}%
\\
\begin{subfigure}[b]{0.45\textwidth}
    \includegraphics[width=\textwidth]{images/samples/10210013_sample.jpg}
    \caption{Low-Resolution Sentinel-2 Satellite Imagery corresponding to the same cluster.}
    \label{fig:sentinel1}
\end{subfigure}
~
\begin{subfigure}[b]{0.45\textwidth}
    \includegraphics[width=\textwidth]{images/samples/10210013.jpeg}
    \caption{Sampled Regions for HR acquisition which are subsequently used for Poverty Prediction.}
    \label{fig:pred1}
\end{subfigure}

\caption{}
\label{fig:sample_visualizations1}
\end{figure*}

\begin{figure*}[!h]
\centering
\begin{subfigure}[b]{0.55\textwidth}
    \includegraphics[width=\textwidth]{images/samples/10210031_all.jpeg}
    \caption{High-Resolution Satellite Imagery (from Digital Globe Systems) corresponding to a cluster.}
    \label{fig:dgs2}
\end{subfigure}%
\\
\begin{subfigure}[b]{0.45\textwidth}
    \includegraphics[width=\textwidth]{images/samples/10210031_sample.jpg}
    \caption{Low-Resolution Sentinel-2 Satellite Imagery corresponding to the same cluster.}
    \label{fig:sentinel2}
\end{subfigure}
~
\begin{subfigure}[b]{0.45\textwidth}
    \includegraphics[width=\textwidth]{images/samples/10210031.jpeg}
    \caption{Sampled Regions for HR acquisition which are subsequently used for Poverty Prediction.}
    \label{fig:pred2}
\end{subfigure}

\caption{}
\label{fig:sample_visualizations2}
\end{figure*}

\begin{figure*}[!h]
\centering
\begin{subfigure}[b]{0.55\textwidth}
    \includegraphics[width=\textwidth]{images/samples/10410002_all.jpeg}
    \caption{High-Resolution Satellite Imagery (from Digital Globe Systems) corresponding to a cluster.}
    \label{fig:dgs3}
\end{subfigure}%
\\
\begin{subfigure}[b]{0.45\textwidth}
    \includegraphics[width=\textwidth]{images/samples/10410002_sample.jpg}
    \caption{Low-Resolution Sentinel-2 Satellite Imagery corresponding to the same cluster.}
    \label{fig:sentinel3}
\end{subfigure}
~
\begin{subfigure}[b]{0.45\textwidth}
    \includegraphics[width=\textwidth]{images/samples/10410002.jpeg}
    \caption{Sampled Regions for HR acquisition which are subsequently used for Poverty Prediction.}
    \label{fig:pred3}
\end{subfigure}

\caption{}
\label{fig:sample_visualizations3}
\end{figure*}

{\small
\bibliography{main.bib}
}